\documentclass[
    superscriptaddress,
    longbibliography,
    twocolumn,
    amsmath,
    amssymb
]{revtex4}

\usepackage{newtxtext}
\usepackage[utf8]{inputenc}
\usepackage[english]{babel}
\usepackage{mleftright}
\usepackage{graphicx}
\usepackage{mathtools}
\usepackage{bm}
\usepackage{xcolor}
\usepackage[version=4]{mhchem}
\usepackage{fancyvrb}
\usepackage{listings}
\usepackage{natbib}
\usepackage{bibunits}
\usepackage{algorithm}
\usepackage{algpseudocode}
\usepackage{amsmath}
\usepackage{xspace}
\usepackage{hyperref}
\hypersetup{hidelinks}
\usepackage{acronym}
\usepackage{comment}
\usepackage{units}
\usepackage{url}

\makeatletter
\def\maketitle{
\@author@finish
\title@column\titleblock@produce
\suppressfloats[t]
}
\makeatother

\defaultbibliographystyle{naturemag}

\algdef{SE}[VARIABLES]{Variables}{EndVariables}
   {\algorithmicvariables}
   {\algorithmicend\ \algorithmicvariables}
\algnewcommand{\algorithmicvariables}{\textbf{Global variables}}

\definecolor{backcolour}{rgb}{0.95,0.95,0.92}
\definecolor{commentgrey}{rgb}{0.4,0.4,0.4}
\definecolor{deepblue}{rgb}{0,0.6,1.0}
\definecolor{lightgrey}{rgb}{0.99,0.99,0.99}
\definecolor{deepred}{rgb}{0.0,0.6,1.0}
\definecolor{deepgreen}{rgb}{1.0,0.0,0.6}

\newcommand{\bc}{\mathbf{c}}
\newcommand{\bw}{\mathbf{w}}
\newcommand{\bx}{\mathbf{x}}
\newcommand{\be}{\mathbf{e}}
\newcommand{\bh}{\mathbf{h}}
\newcommand{\br}{\mathbf{r}}
\newcommand{\bs}{\mathbf{s}}
\newcommand{\bt}{\mathbf{t}}

\newcommand{\bW}{\mathbf{W}}

\newcommand{\bv}{\mathbf{v}}
\newcommand{\bu}{\mathbf{u}}

\newcommand{\voc}{\mathcal{V}}

\newcommand{\loss}{\mathcal{L}}
\newcommand{\dd}{\mathrm{D}}

\newcommand{\linear}{\mathrm{Linear}}

\newcommand{\zero}{{(0)}}
\newcommand{\one}{{(1)}}
\newcommand{\two}{{(2)}}
\newcommand{\none}{{(n)}}
\newcommand{\nzero}{{(n-1)}}
\newcommand{\kone}{{(k)}}
\newcommand{\kzero}{{(k-1)}}
\newcommand{\elu}{\mathrm{ELU}}
\newcommand{\relu}{\mathrm{ReLU}}
\newcommand{\lrelu}{\mathrm{LeakyReLU}}
\newcommand{\gat}{\mathrm{GA}}
\newcommand{\hgat}{\mathrm{HGA}}
\newcommand{\tat}{\mathrm{TA}}
\newcommand{\gru}{\mathrm{GRU}}
\newcommand{\lnorm}{\mathrm{LayerNorm}\xspace}
\newcommand{\splus}{\mathrm{SoftPlus}}
\newcommand{\sigmoid}{\mathrm{Sigmoid}}
\newcommand{\encoder}{\mathrm{AtomEncoder}\xspace}
\newcommand{\transfer}{\mathrm{ResTrans}}
\newcommand{\xenv}{\bx_\mathrm{env}}
\newcommand{\reduce}{\mathrm{Reduce}\xspace}
\newcommand{\angstrom}{\mbox{\normalfont\AA}}
\newcommand{\iiff}{\mathrm{if}\ }
\newcommand{\eelse}{\mathrm{else}}

\newcommand{\modp}{$G_{p}$\xspace}
\newcommand{\modpq}{$G_{pq}$\xspace}
\newcommand{\modpqalt}{$G_{pq}'$\xspace}
\newcommand{\modpqr}{$G_{pqr}$\xspace}

\newcommand{\silabel}{Supplementary Information\xspace}

\begin{document}

\title{3D pride without 2D prejudice: Bias-controlled multi-level\\ generative models for structure-based ligand design}

\author{Lucian Chan}
\affiliation{Astex Pharmaceuticals, Cambridge, UK}

\author{Rajendra Kumar}
\affiliation{Astex Pharmaceuticals, Cambridge, UK}

\author{Marcel Verdonk}
\affiliation{Astex Pharmaceuticals, Cambridge, UK}

\author{Carl Poelking}
\email{carl.poelking@astx.com}
\affiliation{Astex Pharmaceuticals, Cambridge, UK}

\begin{abstract}
Generative models for structure-based molecular design hold significant promise for drug discovery, with the potential to speed up the hit-to-lead development cycle, while improving the quality of drug candidates and reducing costs. Data sparsity and bias are, however, two main roadblocks to the development of 3D-aware models. Here we propose a first-in-kind training protocol based on multi-level contrastive learning for improved bias control and data efficiency. The framework leverages the large data resources available for 2D generative modelling with datasets of ligand-protein complexes. The result are hierarchical generative models that are topologically unbiased, explainable and customizable. We show how, by deconvolving the generative posterior into chemical, topological and structural context factors, we not only avoid common pitfalls in the design and evaluation of generative models, but furthermore gain detailed insight into the generative process itself. This improved transparency significantly aids method development, besides allowing fine-grained control over novelty vs familiarity.
\end{abstract}

\maketitle

\begin{figure*}[hbt]
\includegraphics[width=1.0\textwidth]{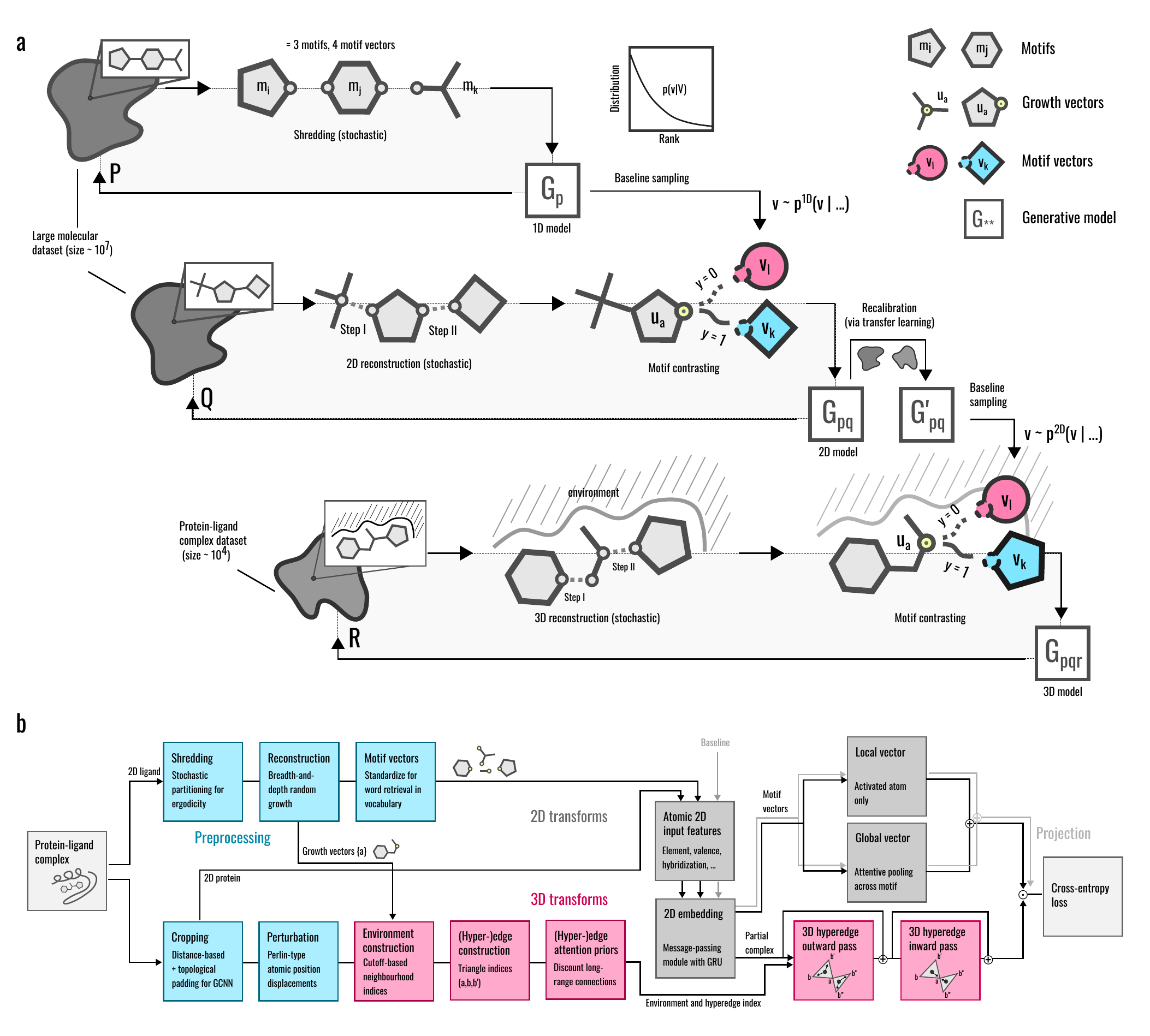}
\caption{ \textbf{Factorized learning framework.} (a) Sequential training pipeline for the $PQR$ model: First, a reference vocabulary is extracted from a molecular library using stochastic shredding rules, resulting in the ``1D'' pseudo-generative model \modp. Second, a 2D generative model is trained on the same molecular library, using contrastive learning, with baseline examples sampled from \modp. Third, the resulting model \modpq is recalibrated on the set of PDB ligands, \modpq $\rightarrow$ \modpqalt. Finally, the 3D  generative model \modpqr is trained with \modpqalt as baseline. (b) Preprocessing, 2D and 3D operations and transforms of the generative architecture during batch training of the network that encodes the 3D context $\bu_{a,3}$. See Methods and the \silabel~for details on data processing and the hypergraph architecture. }
\label{fig:architecture}
\end{figure*}

\begin{bibunit}

Two central questions in drug discovery are: What compound to make next~\cite{schneider_automating_2018}? And -- how~\cite{bostrom_expanding_2018,Blakemore18}? 

In structure-based drug design~\cite{erlanson_twenty_2016,anderson_process_2003}, the decision what to make is guided by structural insight into a protein system of interest: the character of known and suspected druggable sites, their roles in protein function, and the structures of and interactions formed by known ligands. The idea is to improve potency and selectivity of a hit or lead compound by elaborating on its chemical structure within a rationally guided design--make--test cycle. This involves uniquely adapting the compound to the 3D structure of a binding site while maintaining a desirable physicochemical and pharmacokinetic profile. As this cycle typically requires significant experimental resources, the hope is that new computational techniques, in particular those based in machine learning (ML) and deep learning~\cite{Vamathevan19,chen18,paul_artificial_2021}, will continue helping to shorten the time of development, save costs and improve the quality of drug candidates. Notably, recent advances in generative modelling for {\it de novo} molecular design open up possibilities to transfer larger and larger portions of the design process into the {\it in-silico} space~\cite{tong_generative_2021,sousa_generative_2021}. 

There are various ways how ML methods approach the design task, with techniques differing regarding training data, objective and architecture~\cite{olivecrona17,gomezbombarelli18,segler18,popova18,godinez22,krenn20,Cross22,chen21,green21,tan_discovery_2022,piticchio_discovery_2021,krishnan_novo_2021,simm_symmetry_2020,gebauer_inverse_2022}. Two important categories are: 2D generative models that do not explicitly model the protein context, and 3D models, that do. In either case, the design process is often formulated as a graph or string generation task. I.e., the molecular structures are represented as a graph or string~\cite{weininger88,krenn20} and encoded into a latent vector. Various sampling or optimisation strategies can then be used to explore chemical space {\it globally} to generate new candidate compounds~\cite{olivecrona17,gomezbombarelli18,segler18,popova18,godinez22}. In a structure-based design context, however, {\it local} exploration is desirable. This can be achieved within a graph translation or reconstruction setting, where functional groups are added to or substituted into a compound at a specific position~\cite{Cross22,chen21,green21}. Structural constraints, important during hit-to-lead and lead optimization, are then imposed easily to conserve a certain scaffold, growth direction, or pharmacophore.

Even though various molecular generative models have been proposed in recent years, it remains unclear how to fairly, and quantitatively evaluate these models. In graph generation, the current best practice is to demonstrate ``enrichment" over a random baseline for several metrics such as drug-likeness (QED), novelty and diversity~\cite{brown19}. In graph editing, the recovery rate is typically used to assess model performance. However, the impact of $1D$ selection bias (the tendency of a generative model to prioritize common over uncommon chemical motifs, independently of the query compound), is often overlooked~\cite{green21}: The result are misaligned objectives and overly optimistic performance estimates, especially when the sampling distribution underlying the data follows a power-law distribution. This is a well known problem in general recommendation systems~\cite{schnabel16}. In 3D molecular design, as an aggravating factor, topological and geometric information are normally encapsulated into a single generative model. Even with 1D bias correction in place, such a 3D model is trained primarily to reproduce 2D topological or synthetic patterns and constraints, rather than to model the 3D context. Not correcting for both frequency and topological bias not only harms transparency and interpretability, but risks collapsing the learned distributions into a small set of high-frequency modes -- thus reducing both chemical coverage and 3D sensitivity.

In this paper we propose a multi-stage contrastive learning framework for structure-based elaboration of ligands that explicitly controls for sampling and topological bias. The result is a sequential multi-level generative model, where individual stages serve as bias-correcting baselines with respect to which higher-level generative submodels are trained and evaluated. Besides allowing for unbiased performance metrics, the framework (i) provides us with a clear recipe how to augment 3D datasets with the large datasets available for training 2D models; (ii) allows us to easily recalibrate the generative model towards different regions of chemical space or molecular profiles; (iii) improves transparency, as the posterior gives insight into whether a certain recommendation was made based on frequency, topology or structure. The latter point makes this method ideally suited in an assisted interactive design context. Integration into a fully automated design cycle is, however, possible.

\begin{figure}[hbt]
\includegraphics[width=1.0\columnwidth]{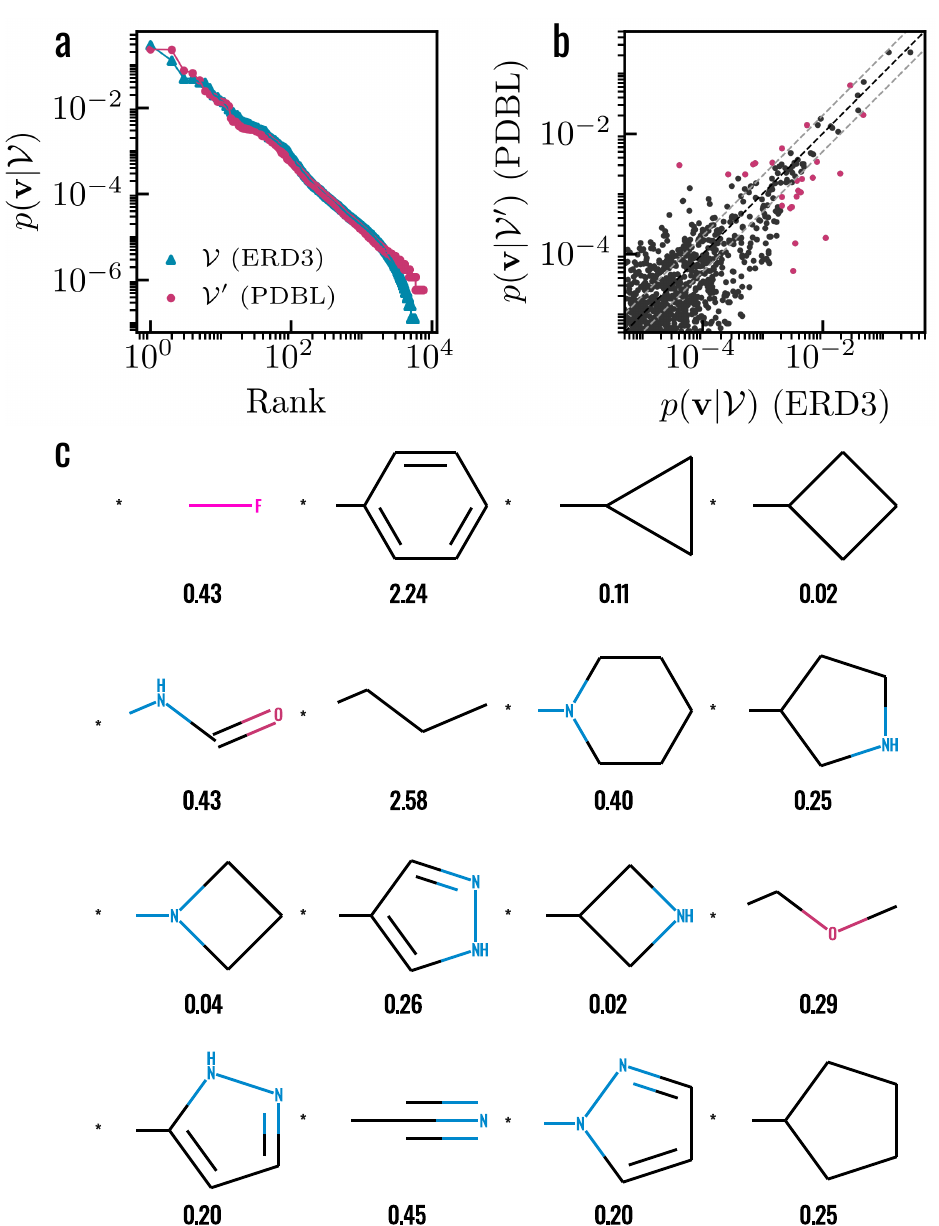}
\caption{ \textbf{ Vocabulary shift between molecular libraries.} (a) Power-law distribution $p(\bv|\voc)$ of the motif vectors for the ERD3 (blue, $\voc$) and PDBL (red, $\voc'$) molecular libraries. (b) Correlation of 1D motif vector probabilities between those two libraries. Highlighted in red are those motif vectors with $\mathrm{max}\{ p(\bv|\voc'), p(\bv|\voc) \} > 0.002$ and probability ratios $p(\bv|\voc)/p(\bv|\voc') > 2$ or $< 0.5$.  The top-sixteen such motifs are shown in panel (c), with their probability ratios $p(\bv|\voc')/p(\bv|\voc)$ (i.e., PDBL over ERD3) indicated below each structure. }
\label{fig:vocabulary}
\end{figure}

\section{Framework~\label{sec:framework}}

We approach the generative objective as a ligand elaboration task: The aim is to construct a model that samples appropriate chemical motifs as functionalizations of an atom-centered growth vector, subject to topological, physicochemical and geometric constraints. When coupled with a conformer generator and applied recursively to its own output, the model can thus grow a starting compound incrementally into a protein environment.

This step-wise elaboration proceeds with respect to a {\it growth vector}, defined here as the atom $a$ of the query molecule to be functionalized. Complementary to this growth vector, a {\it motif vector} represents a chain or ring moiety $k$ together with the atom on that moiety to be attached covalently to $a$. We encapsulate both the growth vector $a$ and motif vector $k$ and their 2D and/or 3D contexts using vector embeddings $\bu_a$ and $\bv_k$, represented by neural networks and to be learnt from data. Given a target chemical space $\mathcal{S}$, the posterior generative distribution can thus be written as $p(\bv_k | \mathcal{S}, \bu_a)$.

The idea is to construct this posterior over chemical motifs $\bv$ of a vocabulary $\voc$ by conditioning independently on (i) the ``1D'' target chemical space $\mathcal{S}$, (ii) the topological ``2D'' context given by the molecular structure of the starting ligand, and (iii) the 3D context of that ligand complexed with a protein target.

To this end, we recast $p(\bv_k \vert \mathcal{S}, \bu_a)$ as a sequential conditional process whereby an initially uniform distribution over chemical motifs $\bv_k$ is refined by ingesting 1D, 2D and 3D information in a step-wise fashion,
\begin{align}
    p(\bv_k | \mathcal{S}, \bu_a) = p(\bv_k | \mathcal{S}) \, q(\bv_k | \bu_{a,2}) \, r(\bv_k | \bu_{a,3}). \label{eq:factorized}
\end{align}
Here we have split the growth vector into a 2D context vector $\bu_{a,2}$ that captures the topology of the ligand independently of its interaction with the protein, and a 3D context vector $\bu_{a,3}$ that depends additionally on the pose of said ligand inside the binding pocket. It is because of this factorization that we refer to the approach as the {\it PQR} framework.

\begin{figure*}[hbt]
\includegraphics[width=1.0\textwidth]{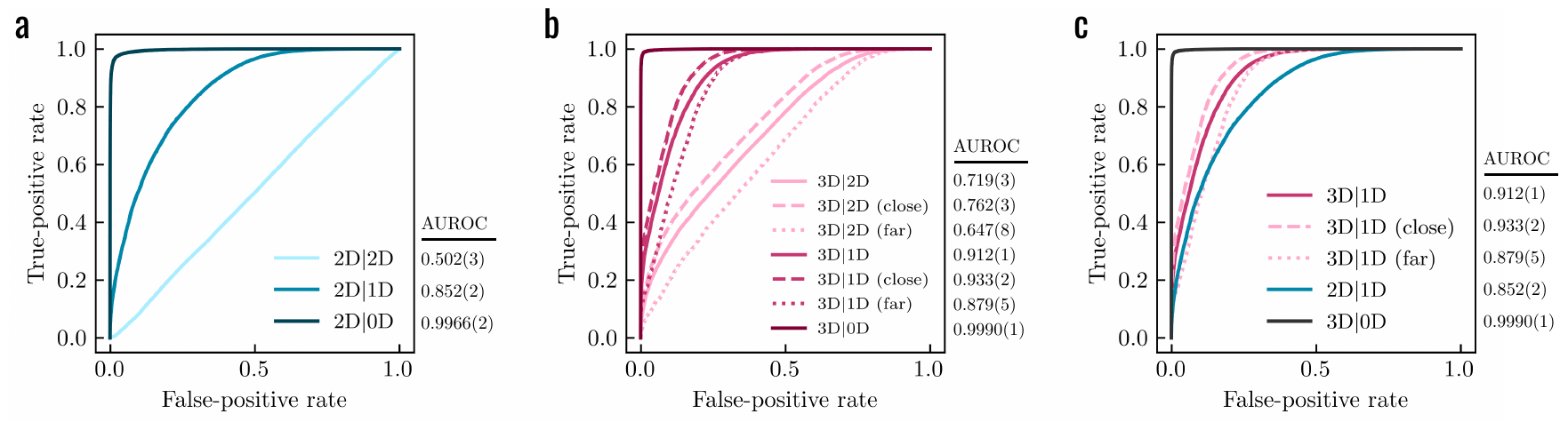}
\caption{ \textbf{Baseline-corrected enrichment in 2D and 3D generative models.} (a) Receiver Operating Characteristics (ROCs) of the 2D model $G_{pq}$ recorded over the set of test ligand reconstructions relative to three baselines: Uniform ``0D'' sampling $G_0$, 1D sampling $G_p$, and 2D sampling $G_{pq}$. Note that the $G_{pq}$ baseline results in an AUC of $0.5$, as by construction. (b) ROCs of the 3D model $G_{pqr}$ recorded over the set of test complex reconstructions against $G_0$ (0D), $G_{p}$ (1D) and $G_{pq}$ (2D) baselines. The dashed and dotted ROCs correspond to a {\it close} and {\it far} subset of the test data, as defined in the main text. (c) Comparison between the 3D (red) and 2D (blue) model evaluated relative to $G_p$, demonstrating the perturbative character of the likelihood factor $r$ in driving late-stage enrichment. }
\label{fig:auroc}
\end{figure*}

As it turns out, this factorized form of the posterior can be constructed term by term (see Methods). The learning protocol necessary to achieve this is summarized in Fig.~\ref{fig:architecture}a, and consists of three key stages: First, the vocabulary $\voc$ and 1D frequency term $p$ are derived via stochastic shredding of a large molecular library (here: the Enamine REAL diverse drug-like set, ERD3), resulting in a pseudo-generative model \modp. Second, the 2D likelihood factor $q$ is constructed using contrastive learning over 2D ligand reconstructions, with \modp serving as a baseline from which ``negative'' examples are sampled. Third, the resulting 2D generative model \modpq serves as the contrastive baseline for the neural network representing the 3D likelihood factor $r$. As during the 2D training stage, this 3D stage is trained on step-wise reconstructions of ligands, but this time into their (experimentally determined) 3D protein environments -- giving rise to an overall generative model \modpqr.

The chemical vocabulary $\voc$, encapsulated by the 1D model \modp, consists of a set of chemical motifs and their relative frequencies. It is a key component of the overall architecture that biases the generation/elaboration towards a drug-like region of chemical space, while maintaining the relative abundance of the distinct chemical groups. The vocabularies differ, however, between the ERD3 library (comprising 25m drug-like compounds) on which \modp and \modpq are trained, and the smaller set of {\it ligand}-like PDB ligands (referred to as PDBL) which feature during the 3D learning stage. This vocabulary shift $\voc \rightarrow \voc'$ from ERD3 to PDBL is visualized in Fig.~\ref{fig:vocabulary} by (a) comparing their power-law distributions over the ranked motifs, as well as (b) by correlating the motif probabilities $p$. As these can easily differ by a factor of two or more (even among common motifs, see panel c), it is important to recalibrate the 2D model \modpq on PDBL before employing it as a 3D baseline (see the transfer learning step in Fig.~\ref{fig:architecture}a). 

This sequential learning framework is architecture- and implementation-agnostic. For this particular study we have resorted to graph and hypergraph convolutional neural networks to represent the 2D and 3D likelihood factors $q$ and $r$, respectively. The 3D training pipeline is summarized in Fig.~\ref{fig:architecture}b, and consists, in particular, of data augmentation steps (stochastic reconstructions, spatially coloured noise), the 3D environment and hyper-edge constructions, 2D and 3D embedding networks, and loss estimation.

Training was performed over the ERD3 library of drug-like molecules, and protein-ligand complexes from the PDB as curated by {\it Binding MOAD}~\cite{moad_1,moad_2,moad_3}. A debiased test set was constructed in line with MOAD's definition of binding-site families. Details on the learning framework, model architecture, implementation, training and test data are provided in the Methods section at the end, as well as the~\silabel.

\begin{figure*}[hp]
\includegraphics[width=1.0\textwidth]{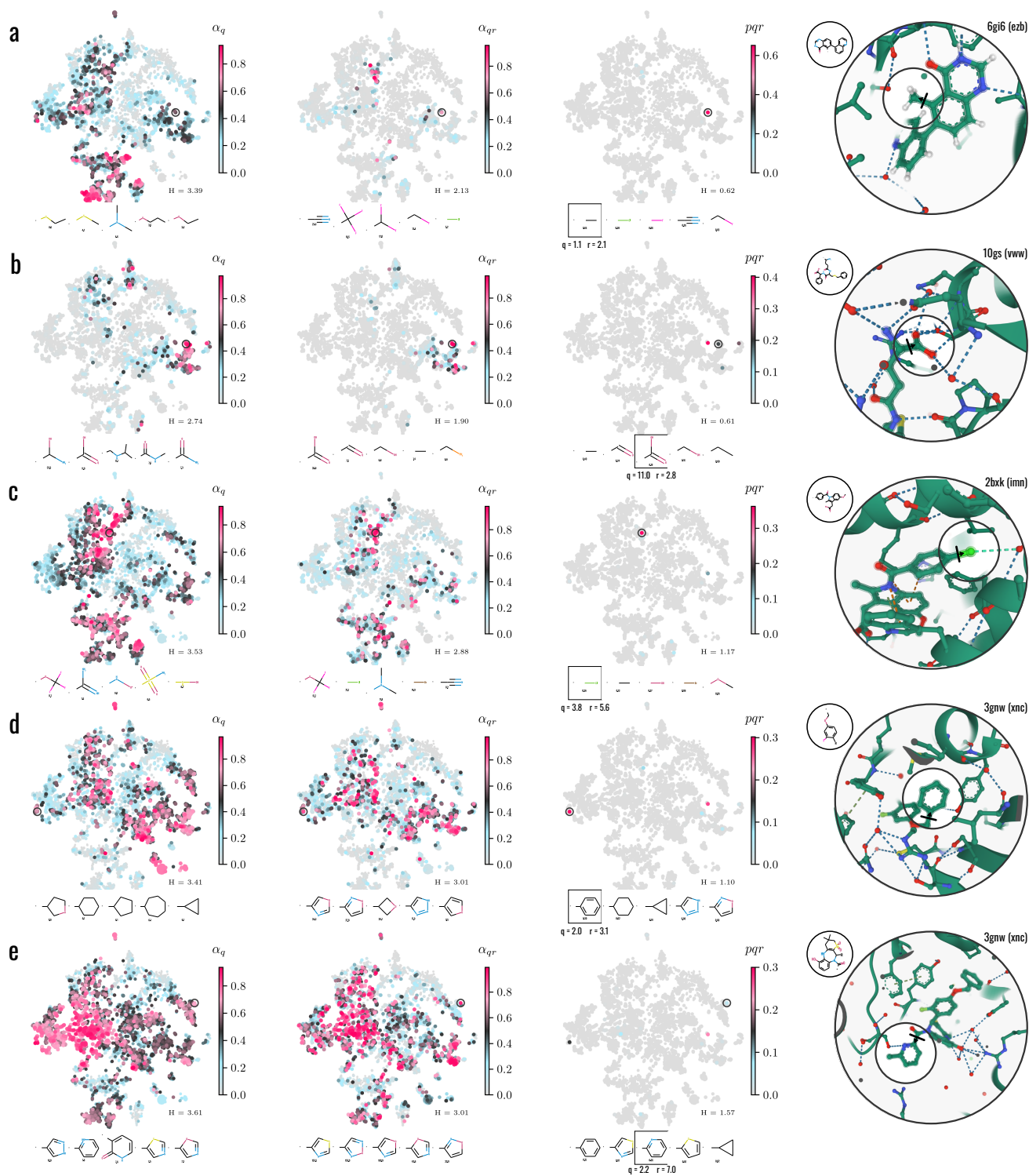}
\caption{ \textbf{Visualization of the posterior.} Likelihood maps over the vocabulary for five example reconstruction steps: The ground-truth motifs are (a) methyl *C, (b) carboxylic acid *COOH, (c) chlorine *Cl, (d) phenyl *C$_6$H$_5$ and (e) pyridine *C$_5$NH$_4$. The maps, derived from stochastic neighbour embedding (t-SNE) over the kernel from Eq.~\ref{eq:2d_kernel}, correspond, from left to right, to the 2D likelihood factor $q$, combined 2D/3D likelihood factor $qr$, and the final posterior $pqr$. The top-five motifs of each model are shown below the likelihood maps, annotated (in small print) with their respective likelihood factor / factor products $q$, $qr$ and $pqr$. In the third column (full posterior) the ground-truth motifs are highlighted with a black outline and labelled with their respective 2D and 3D context factors $q$ and $r$. Note that, to improve contrast, the colour scale for $q$ and $qr$ is based not on the likelihood factors themselves, but their scores $\alpha_q = q/(1+q)$ and $\alpha_{qr} = qr/(1+qr)$. The normalised entropies $H$ of the posteriors are indicated in the bottom right corner of each map (see the \silabel, Sec.~\ref{sec:si_entropy} for further discussion). PDB and ligand HET codes are indicated next to each 3D rendering. In examples (a) and (b) the ground truth is recovered primarily due to the 1D abundance $p$ and 2D propensity $q$, and we therefore cannot credit the 3D model with retrieving the ground-truth structures. 3D structural information is, however, key in recovering the chlorine and pyridine motifs in (c) and (e). }
\label{fig:maps}
\end{figure*}

\section{Results \label{sec:results}}

We will illustrate bias correction, performance and model behaviour within the PQR framework, aiming initially for large-scale statistical validation of the approach, before discussing specific examples and re-tracing experimentally derived structure-activity-relationships.

\subsection{Bias-corrected enrichment}

We have developed the PQR model as a conditional process whereby an initially uniform distribution over chemical space is shaped step-by-step into a context-dependent posterior. As a key benefit of this approach, the performance can be evaluated throughout the sequence by measuring enrichment over the respective baselines.

Appropriate baseline selection is paramount in preventing inflated and uninformative metrics. For illustration, we first estimate the enrichment produced by the pseudo-generative $G_p$ (which considers only the frequency term $p$ of Eq.~\ref{eq:factorized}) over a uniform ``0D'' prior $G_0$ (which weights all motifs with equal probability). This na\"ive model routinely replicates common substituents such as methyl, phenyl, or fluoro-groups -- achieving a top-8 accuracy of $48.7\%$ for reconstructions in the PDBL set. This increases further to $69.1\%$ if, instead of sampling from $p(\bv_k|\mathcal{S})$, we statically rank the motifs by frequency, thus proposing methyl in first place, oxygen in second place, etc. Raw accuracy is thus of limited value when assessing the performance of motif-based generative models for molecular design~\cite{green21}.

To measure the gain due to context information, we will instead rely on AUCs -- Area Under the receiver operating Characteristic (ROC) -- evaluated over the set of test reconstructions: Namely, we record a series of ROC curves, each measuring the performance of a particular model ($G_p \equiv 1\dd$, $G_{pq} \equiv 2\dd$, $G_{pqr} \equiv 3\dd$) over a specific baseline ($G_0\equiv 0\dd$, $G_p$, $G_{pq}$). For the 2D model $G_{pq}$, the resulting ROC curves are shown in Fig.~\ref{fig:auroc}a. First it can be verified that the enrichment that $G_{pq}$ achieves over $G_{pq}$ as baseline is indeed zero (see light blue trend denoted by 2D$|$2D, with an AUC of $0.5$). Replacing the baseline with $G_0$, by contrast, results in a dramatic increase of the AUC to above $0.99$. This drastic improvement is, however, specious, and caused almost entirely by 1D bias (the trivial ability of a model to prioritize common motifs). Substituting $G_0$ with a $G_p$ baseline corrects for this bias, lowering the AUC to $0.85$ (2D$|$1D). Given the large amounts of data that the $G_{pq}$ model is trained on, we speculate that this value is close to optimal and reflects the intrinsic ambiguity among chemically possible functionalisations.

We next determine the performance gain due to 3D information (Fig.~\ref{fig:auroc}b). As a result of inflation, the improvement in AUC over a uniform baseline $G_0$ to above $0.999$ is barely noticeable. The more informative metric is the performance of $G_{pqr}$ over $G_p$ (3D$|$1D), which increases to $0.91$. However: The key measure here is the gain over $G_{pq}$ (3D$|$2D), with an adjusted AUC of $0.719$, up from its 2D value of $0.5$. This 3D$|$2D margin is invaluable, as it allows us to fairly estimate and compare the 3D model performance under different 3D architectures and representations, and at constant 2D performance.

Further insight is gained by investigating separately a {\it close} set of reconstructions that incorporates motif vectors that form a contact of length $\leq 3.5 \unit{\angstrom}$ with at least one heavy atom of the protein; and a {\it far} set where no contacts closer than $4.5 \unit{\angstrom}$ are formed. The AUC drops significantly down to $0.647$ for the {\it far} set, and increases to $0.762$ for the {\it close} set. This indicates how the 3D model anticipates formation of specific interactions, while still capturing lipophilic trends (and, potentially, water-mediated interactions) relevant within the {\it far} set.

Direct comparison between the 3D$|$1D and 2D$|$1D ROC curves (Fig.~\ref{fig:auroc}c) illustrates that the 3D likelihood factors can be thought of as perturbative corrections to the 2D likelihood that improve late-stage rather than early-stage enrichment. This observation is indicative of the tight constraints (large $q$) imposed by the 2D context, and further justifies why decoupling 2D and 3D information is a reasonable approach.

\begin{figure*}[hbt]
\includegraphics[width=1.0\textwidth]{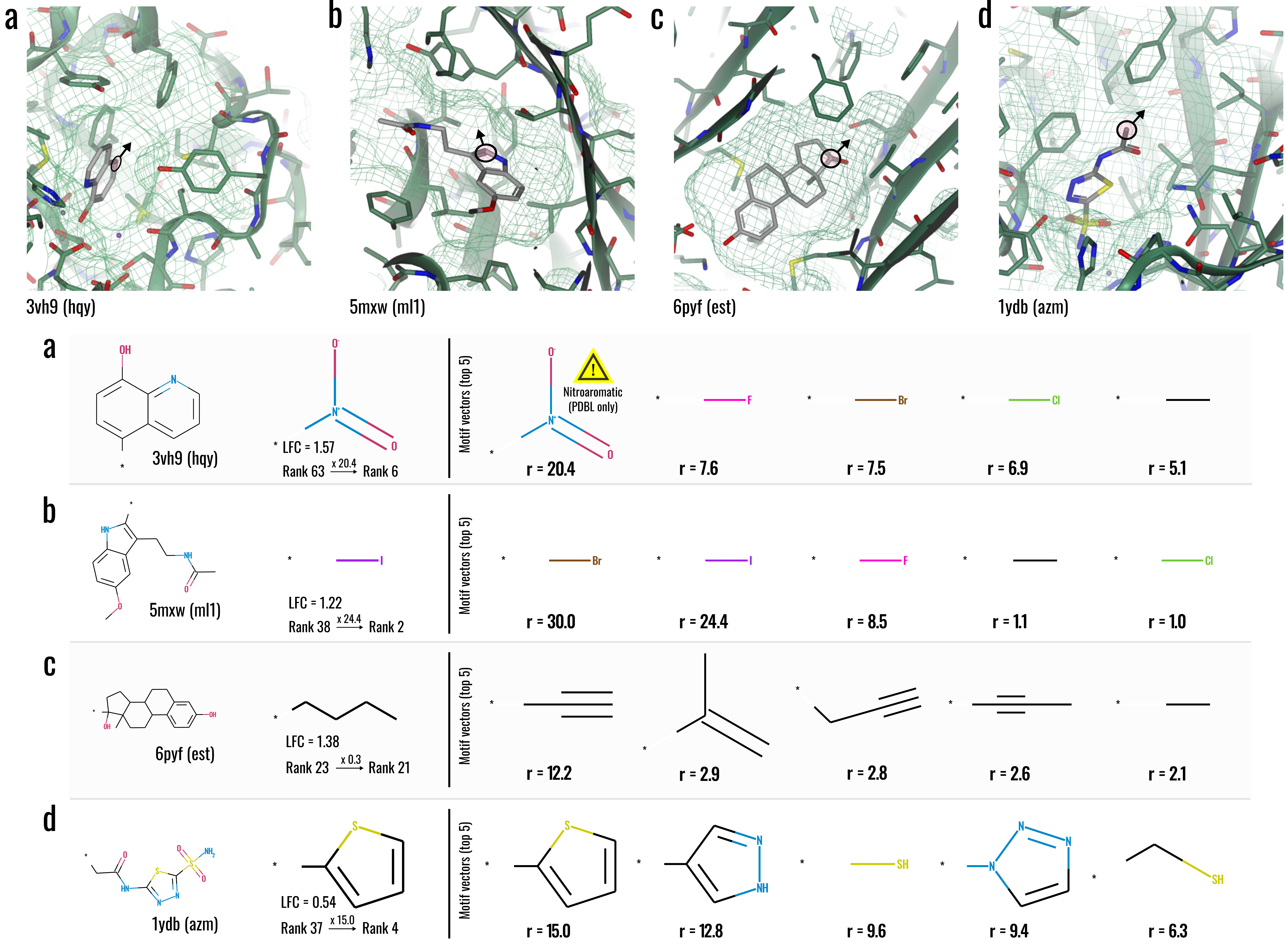}
\caption{ \textbf{Analysis of experimental SARs.} Experimental structure-activity relationships and predicted 3D likelihood factors $r$ for complexes of (a) aeromonas proteolytica aminopeptidase~\cite{example_3vh9}, (b) yellow lupin LLPR-10.2B protein~\cite{example_5mxw}, (c) sex-hormone-binding globulin (SHBG) mutant E176K~\cite{example_6pyf} and (d) carbonic anhydrase (hCA II)~\cite{example_1ydb}. The PDB input structure is identified below each snapshot via its PDB and HET ligand code. In each row, the elaborated ligand and ground-truth motif are displayed on the left-hand side. The ground-truth motifs are annotated with the experimental log-fold change (LFC) in potency, the progression in rank from the 2D model $G_{pq}$ to the 3D model $G_{pqr}$ (rank $R_1 \rightarrow R_2$), with the corresponding 3D likelihood factor indicated above the arrow. On the right-hand side, the top-five predicted motifs are shown, ranked by their likelihood factors $r$, with a top-100 high-pass filter over $pq$. Note that the LFC of $1.57$ in example (a) corresponds to the SAR for hMetAP2, for which the PDB structure 3vh9 here acts as a template; the associated nitroaromatic motif, albeit acceptable as a non-systemic antibacterial agent, is hepatotoxic in humans and represented only in the PDBL vocabulary. In the PQR framework, these toxic groups are automatically discarded when using a drug-like submodel as the 2D component. }
\label{fig:examples}
\end{figure*}

\subsection{Visualization of the posterior}

We demonstrate the transparency of the PQR framework by visualizing the posterior distribution over chemical space. To this end, we reshuffle the conditional process from P-Q-R to Q-R-P: i.e., we first condition on the 2D, then 3D context, and finally on the vocabulary, in order to prevent 1D bias from obscuring the context signal. 

Visualization is achieved with a low-dimensional projection of the vocabulary, derived from the cross-correlation matrix of the likelihood factors (see Methods). Fig.~\ref{fig:maps}a-e depicts the resulting chemical maps for five motif reconstructions. The ground-truth motifs are, from top to bottom: methyl, carboxylic acid, chlorine, phenyl and a (2-substituted) pyridine. The maps in each row correspond to: the 2D likelihood factor $q$ (left), the combined 2D/3D likelihood $qr$ (center), and the full posterior $pqr$ (right).

Note the tight constraints that the 2D context places on the distribution, with large portions of the vocabulary eliminated at the first stage -- as is most pronounced for the methyl and carboxylic acid examples, Fig.~\ref{fig:maps}a,b. Adding the 3D factor silences regions that are incompatible due to sterics or chemical affinity, and redistributes the probability mass to new hotspots: See, e.g., the formation of a lipophilic cluster of small halogen-containing motifs in panel (b), centre. 

The 1D term $p$ results in a strongly narrowed overall distribution $pqr$, with only a handful of high-probability motifs remaining. This slimming of the posterior may be problematic if novelty is desired. Inspecting the $q$ and $r$ likelihood factors independently of $p$ is then a useful workaround -- and easily done within this factorized approach. Following this strategy for the examples at hand gives several interesting insights. First, in (a), the overall model indeed suggested methyl as the most likely motif (see the top-five structures below each chemical map) -- even though $q$ and $r$ evaluate to only $1.1$ and $2.1$, respectively. This means that given the full 2D/3D context, the likelihood of the ground truth just about doubles: a rather modest amplification, illustrating that methyl comes out top simply due to its abundance. Other lipophilic groups, in particular trifluoromethyl and chlorine, achieve much stronger amplification at $qr=18.5$ and $5.2$, respectively.

In (b), the 2D factor ($q=11.0$) already propels the carboxylic acid into second place -- reflecting an affinity to develop amine end groups into glycine. The 3D component further amplifies this motif ($r=2.8$), resulting in a top rank that is ultimately undone by the 1D term. Nevertheless, the two top options (methyl and aldehyde) proposed by the full model could still be converted into the ground truth in subsequent sampling steps and are therefore consistent with the experimental observation.

Summarizing these first two examples, we stress that the apparent success of the model in identifying the ``correct'' motif was, for methyl, driven primarily by its 1D abundance, for carboxylic acid, by its 2D propensity. Hence, even though the 3D factor resulted in an amplification $r>1$ in both cases, we cannot, or at least not to first order, credit the 3D component with recovering these motifs.

The remaining three examples are therefore distinct as, in their case, the 3D factor exceeds the 2D term, $r > q$, with noticeable amplifications of $r=5.6$ (c), $3.1$ (d) and $7.0$ (e). Still, the fact that chlorine (c) and phenyl (d) appear top is again owed to their abundance, with the QR submodel prioritizing trifluoromethoxy (c) and oxazole (d) instead.

\subsection{Further examples}

The PQR model developed in this paper is trained on reconstructions of ligands into their binding sites: Specifically, the model learns to ``fill space`` by maximizing the likelihood of the data -- not necessarily by optimizing the interactions with the protein. In other words, it is designed to prioritize probability over affinity.

And yet: probability and affinity should be expected to correlate, as ligands are implicitly optimized to match the shape and character of their pockets. Here we explore, using a set of experimentally observed structure-activity relationships (SARs), how the model behaves in recovering less common motifs that have been found to significantly improve the potency of a ligand.

To this end, Fig.~\ref{fig:examples} shows examples of functionalizations that resulted in a log-fold change (LFC) in potency of between $0.54$ and $1.57$ (a significant boost in affinity). These examples can be mapped onto known PDB complexes to serve as input structures. We note, however, that none of the PDB complexes of the corresponding binding-site families feature motif or growth vectors similar to those shown in Fig.~\ref{fig:examples}. It is therefore unlikely that the model would be able to base its predictions on any close analogues encountered during training.

The nitroaromatic motif in example (a: aminopeptidase) of Fig.~\ref{fig:examples} is uncommon in drugs due to its hepatotoxic effects, but can be acceptable in non-systemic antibacterial agents. This motif, as well as the iodo substituent in example (b: yellow lupin), are both rare in the PDBL set, with 1D frequencies $p < 0.001$ -- thus rendering recovery by generative methods challenging. And yet: the 3D likelihood factors $r$ amount to $20.4$ for the nitro group, and $24.4$ for iodo, pushing these motifs into first and second place regarding the R submodel, and into sixth and second place with respect to the full PQR model (see annotations below their chemical structures). Interestingly, in (a), the nitro group is succeeded by lipophilic groups F, Br and Cl, in contrast to the polar nitro motif identified in the SAR~\cite{example_3vh9}. Encouragingly, a chloro motif added in an identical position was in a later study found to result in an LFC of $1.36$~\cite{example_3vh9_alt}, lending weight to this particular suggestion.

As another encouraging observation, introduction of both bromine or iodine into the structure of the yellow lupin complex in example (b) would result in a strong clash with a valine group that is contacting the carbon of the growth vector at a distance of $3.8\unit{\angstrom}$. Still, despite this apparent lack of space, the model prioritizes Br and I over the much smaller F and Cl halogens. We speculate that this is the result of the spatially coloured noise that we apply to the structural input during training, which imparts a notion of flexibility onto the 3D model.

Unsurprisingly, the model does not always agree with the experimental findings: In example (c: SHBG) of Fig.~\ref{fig:examples}, the R submodel fails to amplify the butyl motif, with $r = 0.3$ falling significantly below parity. The model does, however, identify several carbyne motifs that, given the linear shape of the cavity in the direction of the growth vector, appear to be an adequate fit -- as can be confirmed using template-based docking (see the \silabel, Sec.~\ref{sec:si_elaboration}).

In the final example (Fig.~\ref{fig:examples}d: hCA II), the ground-truth thiophene is top-rated at $r=15.0$, which helps it progress to 4th rank overall. The validity of the pyrazole, triazole and thiol motifs that follow the top-ranked thiophene with $r > 6$ is challenging to confirm computationally. However, there are several precedents for insertion of triazole and thiol into the local neighbourhood of this growth vector -- albeit in relation to other ligand scaffolds and protein isoforms (see the \silabel, Sec.~\ref{sec:si_elaboration}).

\section{Conclusions}

We introduced a structure-based generative approach to {\it in-situ} ligand elaboration based on a sequential contrastive learning framework. Rigorously decoupling the 1D chemical and 2D topological context from 3D structural information, the framework provides unbiased performance estimates, improved transparency and fine-grained control over the multi-modal posterior from which suitable chemical motifs are sampled. Its modular design efficiently combines the large data resources available for 2D generative modelling with the in comparison sparse datasets of 3D ligand-protein complexes.

Using ligand reconstruction and elaboration examples, we demonstrated how chemical, topological and structural likelihood factors quantify the context-specific amplification of chemical motifs -- thus allowing us to rate functionalisations of an atom-centered growth vector accessibly and transparently. This opens up the possibility of emphasizing novelty, by rescaling or even omitting a 1D bias term that the model incorporates explicitly. The resulting factorized form of the posterior is attractive because it allows evaluating independently the ability of the model to digest 2D and 3D information while preventing inflated performance estimates.

We believe that the transparent design of the generative architecture can considerably benefit structure-based design applications, as well as steer us towards increased automation of the design process. As the learning framework provides feedback on the performance of the 3D component, we hope that future work will quickly surpass the hypergraph approach pursued here, and guide us towards increasingly powerful representations of the structural context. Such improved representations that accurately anticipate interactions with the protein and/or implicit water network would be an important step towards incorporating ligand efficiency and binding affinity objectives into the generative process.\\

{\footnotesize

\section*{Methods}\label{sec:methods}

See below for details regarding: the derivation of the PQR approach (subsection {\it Framework}); the architecture and shredding logic used in this work to implement the framework (subsection {\it Implementation}); the data and training procedure (subsection {\it Training}).

\subsubsection*{Framework}

\textbf{Contrastive construction of the posterior.}
To infer the context-dependent likelihood factors $q(\bv_k | \bu_{a,2})$ and $r(\bv_k | \bu_{a,3})$, our approach is to train separate 1D, 2D and 3D generative models $G_p$, $G_{pq}$ and $G_{pqr}$ back-to-back in a manner that offsets the objective of the ``$n$D'' model against its ``$(n-1)$D'' baseline in a principled way. Using a factorization of the posterior over the chemical vocabulary as posited in Eq.~\ref{eq:factorized}, these three generative models are represented by the distributions
\begin{align}
    p^{1\dd}(\bv_i | \voc) &= p(\bv_i \vert \mathcal{S}) \coloneqq p^{1\dd}_{i \vert a}, \nonumber \\
    p^{2\dd}(\bv_i | \voc, \bu_{a,2}) &= p^{1\dd}(\bv_i | \voc) \times q(\bv_i | \bu_{a,2}), \nonumber \\
    p^{3\dd}(\bv_i | \voc, \bu_{a,2}, \bu_{a,3}) &= p^{2\dd} (\bv_i | \voc, \bu_{a,2}) \times r(\bv_i | \bu_{a,3}), \nonumber
\end{align}
where we have identified the target space $\mathcal{S}$ with the vocabulary $\voc$, and introduced $p^{n\dd}_{i \vert a}$ as a short-hand for the $n$D generative distribution given a growth vector $a$.

We identify the 1D generative distribution with the frequency ratios of the motif vectors $\bv$ measured over a large number of reconstructions:
\begin{align}
    p^{1\dd}(\bv \vert \voc) = \frac{f(\bv_i)}{\sum_j f(\bv_j) }. \nonumber
\end{align}
This distribution is strongly non-uniform as the relative frequencies differ drastically among the motifs: See the discussion around Fig.~\ref{fig:vocabulary}a, which compares $p^{1\dd}$ between two reference sets of molecules (the drug-like ERD3 and the ligand-like PDBL). Note how, in both cases, the eight most common motifs already account for almost 70\% of the total probability mass.

The likelihood factors $q$ and $r$ can be derived sequentially by training estimators on the contrast between a target and baseline distribution. The procedure was depicted schematically in Fig.~\ref{fig:architecture}a. First we define scalar functions 
\begin{align}
    \alpha_n(\bv_i \in \mathbb{R}^d, \bu_{a,n} \in \mathbb{R}^d) \rightarrow [0,1], \ \ n \in \{2, 3\}, \nonumber
\end{align}
represented by neural networks, that resemble projections of motif vectors $\bv_i$ onto growth vectors $\bu_{a,n}$ within the $n\dd$ context. For every reconstruction example, $\alpha_n$ is trained on labels $y_{i|a} \in \{0, 1\}$, assigned so that $y_{i|a} = 1$ for the ground-truth motif $i$, and $y_{i|a} = 0$ for motifs sampled from the $(n-1)\dd$ baseline distribution. To connect the $\alpha_n$'s to the likelihood factors $q$ and $r$, it is convenient to use a binary cross-entropy loss term,
\begin{align}
    \loss_n(\bv_i \vert \bu_{a,n}) &= - y_{i \vert a} \log \alpha_n(\bv_i; \bu_{a,n}) \nonumber \\
    & - (1 - y_{i \vert a}) \log \left[ 1 - \alpha_n(\bv_i; \bu_{a,n})\right], \nonumber
\end{align}
which, over the course of a long training run, is smoothened into a rate differential
\begin{align}
    \langle \loss_n(\bv_i \vert \bu_{a,n}) \rangle_{\mathrm{data}} & \propto - r_{i \vert a} \log \alpha_n(\bv_i; \bu_{a,n}) \nonumber \\
    & - \bar{r}_{i \vert a} \log \left[ 1- \alpha_n(\bv_i; \bu_{a,n}) \right]. \nonumber
\end{align}
By construction, we identify the ``positive'' and ``negative'' rates $r$ and $\bar{r}$ with the target and baseline likelihoods, respectively:
\begin{align}
    r_{i \vert a} &= p^{n\dd}_{i \vert a}, \nonumber \\
    \bar{r}_{i \vert a} &= \bar{p}^{(n-1)\dd}_{i \vert a}. \nonumber
\end{align}
Gradient descent then results in the following relationship between the projections $\alpha_n$ and the likelihood factors:
\begin{align}
    & \partial \langle \loss_n(\bv_i | \bu_{a,n}) \rangle_{\mathrm{data}} = 0 \nonumber \\
    \Leftrightarrow \  & \frac{p_{i|a}^{n\dd}}{p_{i|a}^{(n-1)\dd}} = \frac{\alpha_n(\bv_i; \bu_{a,n})}{1 - \alpha_n(\bv_i; \bu_{a,n})}. \nonumber
\end{align}
Using the factorization of $p^{n\dd}$ and $p^{(n-1)\dd}$ into $p$, $q$ and $r$, we obtain
\begin{align}
    q(\bv_i | \bu_{a,2}) &= \frac{\alpha_2(\bv_i; \bu_{a,2})}{1 - \alpha_2(\bv_i; \bu_{a,2})}, \nonumber \\
    r(\bv_i | \bu_{a,3}) &= \frac{\alpha_3(\bv_i; \bu_{a,3})}{1 - \alpha_3(\bv_i; \bu_{a,3})}. \nonumber
\end{align}
Sampling from the overall generative model is accomplished via discrete inverse-transform sampling over $\bv \in \voc$ with probabilities 
\begin{align}
    p^{3\dd}&(\bv_i | \voc, \bu_{a,2}, \bu_{a,3}) \propto \nonumber \\
    & \frac{f(\bv_i)}{\sum_j f(\bv_j) } \times \frac{\alpha2(\bv_i; \bu_{a,2})}{1 - \alpha2(\bv_i; \bu_{a,2})} \times \frac{\alpha3(\bv_i; \bu_{a,3})}{1 - \alpha3(\bv_i; \bu_{a,3})}. \nonumber
\end{align}
We note that, depending on the application, it may be convenient to omit a subset of the factors of this posterior. The resulting submodels no longer reproduce strictly the molecular space they were trained on, but instead let us emphasize novelty or focus on 3D constraints only (which may be attractive for single-step optimization). One such submodel is the $QR$ (as opposed to $PQR$) model, which omits the 1D bias term, but leaves $q$ and $r$ unaltered. Another option is to apply thresholding to the 1D and 2D factors, for example in the form of a heaviside function that suppresses all motifs with probability $pq$ smaller than some cutoff -- an approach that would thus strike a compromise between frequency, topological and structural constraints.\\

\textbf{Domain recalibration.}
The overall posterior $p^{3\dd}$ as well as the intermediates $p^{1\dd}$ and $p^{2\dd}$ are constructed without the need for ``true'' negative data: ``decoy'' motifs that, upon addition, would prevent the ligand-protein complex from forming  (or, more precisely, decrease the ligand's affinity to below some threshold). For the above construction to work, we do however need to ensure consistency in the chosen baselines across the entire pipeline. The shredding and reconstruction policies applied during the 1D, 2D and 3D phases must therefore be identical. Furthermore, the vocabulary must remain the same during the 2D and 3D cycles. This latter point is problematic, as the ligands appearing in the PDB complexes (PDBL) are distinct from the drug-like space (ERD3) on which the 2D model is trained. The impact on the vocabulary was shown in Fig.~\ref{fig:vocabulary}b-c, which indicated how some rings are acutely more common in ERD3 than PDBL, with frequency ratios of $1:10$ and smaller. 

Training a 2D model exclusively on the PDBL space is not a good approach to bridge this frequency gap, as the number of unique ligands is too small (amounting to ca$.$ 17,000) to adequately train a performant and robust generator. A straightforward solution to this issue is to recalibrate the 2D model before using it as a 3D baseline in a way that corrects for the phenotypic difference between ERD3 (with vocabulary $\voc$) and PDBL (with vocabulary $\voc'$). This is achieved with a brief transfer-learning step inserted between the 2D and 3D phase which performs the necessary adjustment (see the transition from $G_2$ to $G_2'$ in Fig.~\ref{fig:architecture}a):
\begin{align}
    p^{1\dd}(\bv_i | \voc) & \rightarrow p^{1\dd}(\bv_i | \voc'), \nonumber \\
    p^{2\dd}(\bv_i | \voc, \bu_{a,2}) & \rightarrow p^{2\dd}(\bv_i | \voc', \bu_{a,2}). \nonumber
\end{align}
We point out that this consistency constraint is needed only during training. In a deployment scenario, the 1D, 2D and 3D components (represented by their likelihood factors $p$, $q$ and $r$) can be treated in a modular fashion. E.g., the motif frequencies can be re-scaled to prioritize novel over standard chemistry, or disallow certain types of motifs. Furthermore, the 2D model can be substituted with a generator specialized towards a particular physicochemical or DMPK profile without the need to re-train the 3D component.\\

\textbf{Normalization of the posterior.} 
Given that the likelihood factors are derived from a contrastive rate differential, there is no guarantee that the posteriors $pq$ and $pqr$ are still normalized over the vocabulary. Let $z^{2\dd}$ and $z^{3\dd}$ be the normalization constants
\begin{align}
    z^{2\dd}_{a} &= \sum_{\bv_i \in \voc} p(\bv_i | \voc) q(\bv_i | \bu_{a,2}), \nonumber \\
    z^{3\dd}_{a} &= \sum_{\bv_i \in \voc} p(\bv_i | \voc) q(\bv_i | \bu_{a,2}) r(\bv_i | \bu_{a,3}). \nonumber
\end{align}
We then define effective likelihood factors $\hat{q}$ and $\hat{r}$ as 
\begin{align}
    \hat{q}(\bv_i | \bu_{a,2}) = \frac{1}{z^{2\dd}_a} q(\bv_i | \bu_{a,2}), \nonumber \\
    \hat{r}(\bv_i | \bu_{a,3}) = \frac{z^{2\dd}_a}{z^{3\dd}_a} r(\bv_i | \bu_{a,3}). \nonumber
\end{align}
The $q$ and $r$ values reported in the main text correspond to these corrected likelihood factors.\\

\textbf{Visualization of the posterior.}
The 2D projections in Fig.~\ref{fig:maps} are based on a kernel $k_{ij}$ defined between any two motif vectors $\bv_i$ and $\bv_j$, derived from the cross-correlation matrix of the 2D network scores $\alpha_2(\bv_i; \bu_{a,2})$, sampled over a large number $N$ of randomly selected growth vectors $a$,
\begin{align}
    k_{ij} = \frac{1}{N} \sum_a \hat{\alpha}_2(\bv_i; \bu_{a,2}) \, \hat{\alpha}_2(\bv_j; \bu_{a,2}), \label{eq:2d_kernel}
\end{align}
where $\hat{\alpha}$ denotes the normalised (whitened) scores. We construct a distance metric $d_{ij}$ from $k_{ij}$ using $d_{ij} = 1 - k_{ij}$ and obtain a 2D representation of the vocabulary by applying t-SNE~\cite{tsne08} to the resulting distance matrix. Projecting the values of the posterior onto this 2D map, we can visualize how the probability density changes as we incorporate 2D, 3D and frequency information.\\

\textbf{Related Work.} 
It is well known that recommendation algorithms suffer from selection bias when trained directly on raw data without appropriate countermeasures. Methods based on inverse propensity weighting have been proposed to alleviate this issue~\cite{schnabel16}, with recent work illustrating how bias correction helps to improve the \textit{fairness} of the algorithms towards under-recommended items and domains. These methods involve estimation of a propensity score, which is typically computationally expensive. 

As an alternative approach, contrastive learning has been applied to bias reduction~\cite{zhou21} with the aim to learn a high-quality representation via a self-supervised pretext task. This technique has been widely applied in various domains, including computer vision~\cite{khac20}, graph data~\cite{you20} and molecular data~\cite{wang22}. With an appropriate proposal distribution for negative sampling, this framework simultaneously improves the discriminative power of the model and reduces exposure bias. Under mild conditions, the contrastive loss can be viewed as a sampling-based approximation of an inverse propensity weighted loss.~\cite{zhou21}

\textit{Fairness} is not the primary objective pursued in this paper, as generative models for molecular design -- as opposed to recommender systems -- are constructed so as to sample from a target distribution without over-emphasizing rare chemistry. Nevertheless, both for local optimization as well as fully automated ML-driven design, fairness with respect to how distinct regions of chemical space are weighted by the generative model provides improved control and transparency, and avoid misaligned objectives during training by incentivizing the model to represent under-explored chemical motifs rather than brush over it.

DeepFrag, developed by Green and co-workers~\cite{green21} is directly related to our work in that it uses reconstructions (in their case, of terminal groups) to learn 3D-compatible modifications to bound ligands. Their 3D grid-based convolutional architecture does not, however, correct for any 1D or 2D bias (which we believe is an essential aspect to any 3D-aware generative model, as discussed in great detail in the Results section). It also remains uncertain to what level their architecture can incorporate 2D chemical constraints, seeing that DeepFrag has been trained on only a very limited number of ligand-like molecules. Furthermore, by representing motifs as static fingerprints within a deterministic architecture, it is unclear how DeepFrag can learn multi-modal posteriors and avoid undesirable chemical cross-talk between structurally related but chemically distinct groups.

\subsubsection*{Implementation}

The PQR framework is general and not designed with a particular architecture in mind. It has a modular structure, so that individual components can be substituted or amended without requiring re-training of the entire sequence. In the implementation, the key components that require specification are the neural networks that encode the 2D and 3D growth vectors $\bu_{a,2}$ and $\bu_{a,3}$, the networks that encode the motif vectors $\bv_{k,2}$ and $\bv_{k,3}$, and the shredding and reconstruction policies that define the vocabulary and training batches.\\

\textbf{Shredding and reconstruction.}
The training strategy is to treat the elaboration as a stochastic reconstruction task: i.e., a bound ligand is first shredded and then regrown step by step into its 3D environment. Both the shredding and reconstruction are treated stochastically in order to guarantee ergodicity of the generative sampling. 

The shredding rules, together with the reference molecular space $\mathcal{S}$ (ERD3 and PDBL) implicitly define the chemical vocabulary and hence the pseudo-generative model \modp. To construct \modp, molecules are sampled from $\mathcal{S}$ and observation counts (frequencies) $f(\bv_k)$ are aggregated for all  motifs $\{ k \}$.

Shredding is performed as follows: First, bonds between ring and non-ring atoms are cut (with the exception of exocyclic oxygen =O to preserve conjugation in aromatic rings). Whereas the resulting ring isolates are added as motifs without additional processing, chains/linkers are further decomposed stochastically. This decomposition is carried out iteratively, at every stage starting from a seed atom sampled with a prior weight $w$ composed of the sum of its bonded degree and explicit valence. The motif is expanded around the seed atom by following covalent links shell-by-shell up to a maximum topological radius sampled uniformly over $[0,2]$. Growth happens, with equal probability, either directionally or isotropically along the bonds. Once the topological radius has been reached, the motif is isolated from the rest of the chain by cutting all peripheral bonds, and the member atoms are removed from the set of eligible seeds.

Reconstructions are generated {\it diffusively} with respect to a seed motif, see the section on {\it Data augmentation} below.\\

\textbf{2D and 3D architectures.} Here we give only a high-level overview of the model structure. Architectural details on the encoder networks, graph and hypergraph attention operators etc$.$ are provided in the \silabel, Sec.~\ref{sec:si_2d}-\ref{sec:si_3d}.

The 1D baseline, which is constructed directly from a reference molecular library and therefore parameter-free, is the na\"ive motif frequency model $G_{p}$. The 2D baseline, $G_{pq}$, which encompasses $G_{p}$, additionally consists of three graph attention networks: a model to identify potential growth vectors, a model to represent the likelihood factor $q$, and a model to predict the bond order between the core structure and motif vector. The 3D model, $G_{pqr}$, which in turn encompasses $G_{pq}$, furthermore incorporates a hypergraph attention network to represent the likelihood factor $r$.\\

On the topological level, we represent both ligands and proteins as graphs with nodes $n$ representing atoms and edges $(n,n')$ representing covalent bonds. Both the networks for $\bu$ and $\bv$ incorporate a 2D message-passing neural network $e_2$ which embeds the atoms of the graphs into 256-dimensional latent-space vectors $\bx_n$. This network is shared across the ligand and the protein. The second stage of the networks then processes these embeddings further into growth and motif vectors $\bu$ and $\bv$, respectively. 

This second step is straightforward for the 2D model $G_{pq}$, and is accomplished with three dense convolutional layers applied to the node embeddings $\bx_n$. For the 3D model $G_{pqr}$, however, the 3D context, consisting of the relative positions of the atoms, needs to be represented as well. Given that we ultimately predict a scalar quantity, this embedding step should be carried out with an $SE(3)$-invariant architecture. Here we opt for a hypergraph approach, where we introduce additional virtual nodes and edges representing triangular relationships $(a,n,n')$. This approach is attractive in that it captures 3D information similar to hard-coded basis-function-based convolutional descriptors such as SOAP~\cite{bartok_representing_2013}, which are themselves too high-dimensional to be practical for neural networks. We construct these hypergraphs locally for the environment of an atom-centered growth vector, using a spherical cutoff $r_c = 7.5 \unit{\angstrom}$ applied to protein atoms and a smaller cutoff $r_c = 3.0 \unit{\angstrom}$ applied to ligand atoms -- the rationale for the latter being that the first two covalent coordination shells around the growth vector are needed to model implicitly a local frame and growth direction.

The hypergraph convolutions consist of two stages: An {\it outward} pass communicates information about a triangle $(a,n,n')$ to the environment atoms $n$ and $n'$; subsequently, an {\it inward} pass transmits information from $n$ and $n'$ back to $a$. Attentive pooling along all triangular incoming and outgoing virtual nodes is carried out using a transformer-type attention mechanism which additionally incorporates attention priors: These priors are designed to ensure smoothness with respect to atoms moving into and out of the neighbourhoods as well as discount ``distant'' over ``close'' triangular connections.

The motif vectors are derived from the 2D embeddings $x_n$ in a fashion similar to how the 2D growth vectors $\bu_{a,2}$  are modelled. However: In addition to a vector that captures the local neighbourhood around the attachment atom, we also construct a global vector by pooling attentively over the entire motif. The final motif vector $\bv$ is then taken to be the sum of the two.

The likelihood scores $\alpha_n(\bv_k; \bu_{a,n})$ for both the 2D and 3D stage are modelled as an activated scalar dot product between the growth and motif vectors,
\begin{align}
    \alpha_n(\bv_i; \bu_{a,n}) = \sigmoid\left( \frac{\bv_{i,n} \cdot \bu_{a,n}}{2 \sqrt{d}} \right). \nonumber
\end{align}
This projection mechanism motivates why the motif vectors (as opposed to only the growth vectors) need to be represented using deep neural networks, as feature and latent-space collisions could otherwise result in cross-contamination among structurally similar but chemically distinct moieties. A rich embedding $\bv_k$ should furthermore allow the model to accurately represent multi-modal posteriors, where disparate chemical groups can couple to the same growth vector.\\

\textbf{Software.}
RDKit~\cite{greg20} was used to read, write and compute atom and bond features of ligands and proteins. The graph attention networks were implemented in PyTorch~\cite{paszke19} and PyTorch Geometric~\cite{fey19}, with custom extensions for efficient hypergraph construction.

\subsubsection*{Training}

\textbf{Training and test data.}
Training within the PQR framework relies on two types of datasets: A (large) set of drug-like compounds from which we derive the 1D frequencies and 2D model $G_{pq}$; and a dataset of ligand-protein complexes, from which we construct the domain-recalibrated 2D model $G_{pq}'$ and 3D model $G_{pqr}$.

For the set of drug-like molecules, we have used 25 million compounds from the {\it Enamine REAL diverse drug-like set} (ERD3)~\cite{enamine_real}, which itself is a subset of the {\it REAL drug-like set} -- constructed such that the Tanimoto similarity calculated over Morgan2 512-bit fingerprints~\cite{rogers10} is less than 0.6 between any two molecules. These compounds satisfy the ``rule of 5''~\cite{shultz_two_2018}, which places upper limits on the molecular weight, surface logP, polar surface area, and the number of hydrogen bond acceptors and donors.

The ligand-protein complexes are taken from the 2019 release of Binding MOAD -- a curated subset of the PDB containing 38702 ligand-protein complexes~\cite{moad_1,moad_2}. This dataset considers only complexes of biologically relevant ligands: small organic molecules, oligopeptides and -nucleotides, but no crystallographic additives, solvents etc$.$ (which the PDB may classify as ligands but that are not interesting in the context of this work). MOAD furthermore splits the ligand complexes into 10500 families based on binding-site similarity as quantified by APoc~\cite{apoc}, with sites of the same family having an association $p$-value of smaller than $0.05$. Here we used the family definitions to construct a test set of 1001 complexes, none of which are part of any family represented in the training set.\\

\textbf{Batch training.} 
The training pipeline for $g_3$ (Fig.~\ref{fig:architecture}b) is structured as follows: Each mini-batch consists of 40 protein-ligand complexes, with each complex giving rise to ca$.$ eight reconstructions on average. The protein and ligand structures are perturbed by spatially coloured noise (see {\it Data augmentation}) and then cropped around the ligand with the same cutoff used to define the size of the local neighbourhoods for the hypergraph convolutions. This cutoff region is subsequently expanded along the bonded network to match the field of view of the 2D message-passing network $e_2$ with respect to the boundary atoms. The bound ligand is shredded and a reconstruction pathway is sampled to reconnect the shreds in sequence. The motif vectors encountered in the reconstruction are normalized to allow for hash-based retrieval in the vocabulary. 

For each reconstruction, the environment associated with the growth vector is identified, and the triangular hyperedges and nodes within that neighbourhood are constructed and featurized. The molecular graph (encompassing the ligand fragments, protein and motifs) are embedded into a 2D latent space, from which subsequently the motif vectors and 3D-aware growth vectors are constructed in the manner described above.

To evaluate the contrastive loss, counter-examples are sampled from the baseline model $G_{pq}$ with the ligand fragments and growth vector atoms as queries. These counter-examples are convoluted identically to the motif vectors of the reconstructions and projected onto the growth-vector encodings. Optimization is based on a binary cross-entropy loss function applied over the projections $\alpha$.

The results reported in the main text are from a training sequence as follows: The 1D baseline $G_{p}$ is derived from 1m compounds sampled randomly from ERD3. The 2D baseline $G_{pq}$ is trained over 40 epochs, each consisting of 250k compounds sampled randomly from ERD3. The recalibration of $G_{pq}$ from ERD3 to PDBL is performed over ten epochs. Finally, the 3D model $G_{pqr}$ is trained over 40 epochs, with each epoch being unique regarding the reconstruction sequence and co\"ordinate perturbations.\\

\textbf{Data augmentation.}
To improve diversity, regularization and robustness, we employ three data augmentation measures during training: Stochastic ligand reconstructions, spatially coloured noise, and small torsion rotations on protein side chains.

Stochastic reconstruction pathways are generated from a motif adjacency matrix. First, a seed motif is selected randomly from the shreds (which are themselves stochastic, see section on {\it Implementation} above). The molecular core is then grown recursively from this seed motif, by following randomly the motif-motif edges accessible from the core, thus adding additional motifs to the core step-by-step.

The atomic co\"ordinates of both the protein and ligand are perturbed with spatially coloured noise generated by interpolating white noise along the bonded network. First, atomic displacements $\Delta_{x,a}^{(0)}, \Delta_{y,a}^{(0)}, \Delta_{z,a}^{(0)}$ are drawn from a normal distribution $\mathcal{N}(0,1)$. Interpolation occurs iteratively with respect to the covalent adjacency matrix $A_{ab}$ , 
\begin{align}
    \Delta_{*,a}^{(n)} \rightarrow \frac{1}{1 + \sum_{b} A_{ab}} \left(\Delta_{*,a}^{(n-1)} + \sum_{b} A_{ab} \Delta_{*,b}^{(n-1)} \right), \nonumber
\end{align}
up to $n=5$. The resulting displacements are, finally, rescaled uniformly to match a target overall standard deviation $\sigma=0.5\unit{\angstrom}$, and clamped at a maximum magnitude of $2\sigma$.

Flexible torsional degrees of freedom $\phi$ of the amino-acid side chains are perturbed uniformly in the interval $\phi \in [-10^\circ,+10^\circ]$.\\


\textbf{Acknowledgements.}
LC acknowledges funding from Astex through the Sustaining Innovation Postdoctoral Program. We thank Chris Murray and Davide Branduardi for thoughtful comments on the manuscript, and Lucy Colwell for fruitful discussions. \\

\textbf{Data availability.}
The datasets used for this study are publicly available, for pointers see \url{https://github.com/capoe/libpqr/-/tree/master/data}.\\

\textbf{Code availability.}
The source code and pre-trained models can be accessed at \url{https://github.com/capoe/libpqr}.\\

}

\end{bibunit}

\clearpage
\setcounter{page}{1}
\setcounter{figure}{0}
\setcounter{section}{0}
\setcounter{equation}{0}
\renewcommand{\thepage}{S\arabic{page}}
\renewcommand{\thesection}{\Alph{section}}
\renewcommand{\thesubsection}{\alph{section}}
\renewcommand{\thetable}{S\arabic{table}}
\renewcommand{\thefigure}{S\arabic{figure}}

\title{SUPPLEMENTARY INFORMATION \\ \vspace{0.5cm} {3D pride without 2D prejudice: Bias-controlled multi-level\\ generative models for structure-based ligand design}}

\maketitle

\begin{bibunit}

Below we provide additional details on the implementation of the framework, in particular regarding the neural network architectures used to represent the 2D and 3D models (Sections~\ref{sec:si_1d}-\ref{sec:si_3d}). We furthermore extend the discussion of how the 1D, 2D and 3D context affects the information content of the posterior  (Section~\ref{sec:si_entropy}). Finally we evaluate, using structural precedents, the realism of several of the suggestions made by the framework in relation to the experimental structure-activity relationships discussed in the main text (Section~\ref{sec:si_elaboration}).\\

\textbf{Notation.} 
The following definitions are used: Vector embeddings are denoted by $\bx$ and $\bh$, which, except for the input feature vectors, are elements of $\mathbb{R}^{n \times d}$ with batch dimension $n$ and hidden dimension $d=256$; $\be$ are edge vectors describing covalent edges used in 2D convolutions; $\bt$ are hypernode vectors representing triangular relationships between atoms in 3D. Indices $a,a',a''$ denote atoms of the query molecule incorporating the growth vector $\bu$. Indices $b,b', \dots$ correspond to atoms of the 3D environment, which could belong either to the query ligand or surrounding protein. Indices $i,j,k$ qualify motif atoms and the motif vector $\bv$. Graph convolution operations are indicated by square brackets $G[\dots]$. Round brackets are used for both point-wise function evaluation as well as row-wise convolutions; $\circ$ denotes point-wise multiplication; $||$ means concatenation of vectors along the non-batch axis. $\bW$ and $\bc$ are learnable parameters of the model; $\gamma$'s are attention weights. 

Nonlinear activation functions are referred to using the terms: ELU (Elastic Linear Unit), ReLU (Rectifying Linear Unit), LeakyReLU (Leaky Rectifying Linear Unit), Sigmoid, SoftPlus. These are defined as follows: 
\begin{align}
    \elu(x) &= \left\{ \begin{array}{cl}
         x & \iiff x > 0, \\
         \exp(x) - 1 & \eelse,
    \end{array}\right.\\
    \relu(x) &= \left\{ \begin{array}{cl}
        x & \iiff x > 0, \\
        0 & \eelse,
    \end{array}\right.\\
    \lrelu(x) &= \left\{ \begin{array}{cl}
        x & \iiff x > 0, \\
        0.01 x & \eelse,
    \end{array} \right.\\
    \sigmoid(x) &= \frac{1}{1 + \exp (-x)}\\
    \splus(x) &= \log(1 + \exp x)
\end{align}

\section{1D model}
\label{sec:si_1d}

The pseudo-generative 1D model $G_p$ is defined by the set of motif vectors $\{ \bv_i \}$ with frequencies $\{ f(\bv_i) \}$ that constitute the vocabulary $\voc$:
\begin{align}
    \bv_i \sim p^{1\dd}(\bv_i | \voc) = \frac{f(\bv_i)}{\sum_j f(\bv_j)}. \label{eq:si_pseudo}
\end{align}
We refer to $G_p$ as a ``1D'' model, because it samples motifs independently of the 2D structure of the query molecule and growth vector, using only a string hash representation of the motif vector.\\

\textbf{Null metrics.}
$G_p$ achieves considerable enrichment over a uniform baseline $G_0$ as illustrated in tables~\ref{Tab:null_auc} and~\ref{Tab:null_acc}. Beyond serving as a contrastive baseline for 2D model training, $G_p$ thus provides us with a set of {\it null} metrics against which 2D and 3D performance should be offset.

\begin{table}[H]
\begin{center}
\begin{tabular}{ c|c } 
              & AUC \\ 
\hline
ERD3  &  $0.9884 \pm 0.0005$ \\ 
PDBL  &  $0.9893 \pm 0.0007$ \\ 
\hline
\end{tabular}
\end{center}
\caption{\textbf{Null enrichment.} AUC measured for the 1D pseudo-generative model $G_p$ over 10000 reconstructions in the ERD3 (Enamine REAL) and PDBL (PDB ligand) sets. The negative baseline examples were drawn from the uniform $G_0$. Note that the AUC is independent of how many ($n \geq 1$) negative baseline samples  are considered per reconstruction.}
\label{Tab:null_auc}
\end{table}

\begin{table}[H]
\begin{center}
\begin{tabular}{ c|c|c|c|c } 
              & Top-$1^{a}$ & Top-$1^{b}$ & Top-$8^{a}$ & Top-$8^{b}$\\ 
\hline
ERD3          & $28.7$ & $10.9 \pm 0.2$ & $63.2$ & $42.4 \pm 0.4$ \\ 
PDBL          & $22.9$ & $11.6 \pm 0.2$ & $69.2$  & $48.7 \pm 0.2$ \\ 
\hline
\end{tabular}
\end{center}
\caption{\textbf{Null accuracy.} Top-$k\%$ accuracy ($k\in {1,8}$) and associated standard error measured for ($a$): a ``top-$k$ motifs`` model that statically ranks motifs based on their frequencies; and ($b$) the pseudo-generative 1D model (Eq.~\ref{eq:si_pseudo}).}
\label{Tab:null_acc}
\end{table}

\section{2D model}
\label{sec:si_2d}

The 2D model consists of three components: a growth-vector prediction stage that samples a growth-vector atom $a$ given the query structure; a motif sampling stage that generates a motif $\bv_i$ given a specific growth vector $\bu_a$; and a bonding model that samples a bond order $b$ (single, double, triple) given $\bv_i$ and $\bu_a$:
\begin{align}
    a &\sim p(a | \{ \bu_{a',2} \}), \\
    \bv_i &\sim \frac{f(\bv_i)}{\sum_j f(\bv_j) } \times \frac{\alpha_2(\bv_i; \bu_{a,2})}{1 - \alpha^{2\dd}(\bv_i; \bu_{a,2})}, \\
    b &\sim p(b | \bv_{i}; \bu_{a,2}).
\end{align}
The growth-vector and bond-order prediction models are outside the scope of this work, and therefore not specified further. Below we will focus on how the likelihood factor score $\alpha_2$ is represented.\\

\textbf{Input features.}
Atoms (nodes) and bonds (edges) of the molecular graph are featurized using a set of chemical properties as summarized in Tab.~\ref{tab:atomfeature} and~\ref{tab:bondfeature}. Categorical features are encoded in a one-hot format, resulting in an overall input dimension of 36 for the atom and 6 for the bond feature vectors.\\

\begin{table}[h]
\centering
    \begin{tabular}{l l c}
        Feature & Description & Dim. \\
        \hline
         Atom type & Element (one-hot) & 12  \\
         Atom degree & Neighbour count (one-hot) & 7 \\
         Radical electrons & Radical-electron count & 1\\
         Formal charge  & Charge in $e$ & 1 \\
         Hybridization & $s$, $sp^{1}$-$sp^{3}$, $sp^{3}d$, $sp^{3}d^{2}$, other  & 7 \\
         Aromaticity & Aromatic indicator & 1 \\
         Hydrogen count & Explicit + implicit & 5 \\
         Chirality & CIP R and S indicators & 2\\
         \hline
    \end{tabular}
    \caption{\textbf{Atomic input features.} Atom node attributes used by all 2D embedding networks.}
    \label{tab:atomfeature}
\end{table}

\begin{table}[h]
    \centering
    \begin{tabular}{l l c}
        Feature & Description & Dim. \\
        \hline
        Bond type & Single -- triple, aromatic, conjugated & 5 \\
        Ring bond & Ring indicator & 1 \\
         \hline
    \end{tabular}
    \caption{\textbf{Bond input features.} Covalent edge attributes used by all 2D embedding networks.}
    \label{tab:bondfeature}
\end{table}

\textbf{Graph architecture.}
Starting from the atom and bond features $\bx$ and $\be$, an $\encoder(\bx, \be)$ builds 2D latent-vector representations using the following set of convolutions:
\begin{align}
    \bx^\zero &= \lrelu\left(\linear\left(\bx\right)\right), \label{eq:atomenc_start} \\
    \bh^\one &= \elu\left(\gat_0\left[\bx^\zero,  \be^\zero \right] \right), \\
    \bx^\one &= \relu\left(\gru\left(\bh^\one, \bx^\zero\right)\right), \\
    \bh^\none &= \elu\left(\gat_1\left[\bx^\nzero\right]\right), \\
    \bx^\none &= \relu\left(\gru\left(\bh^\none, \bx^\nzero\right)\right), \\
    \bx_{2\dd} &= \lnorm\left(\bx^{(4)}\right). \label{eq:atomenc_end}
\end{align}
Here $\linear(\bx) = \bW \bx + \mathbf{b}$ is a dense linear layer with weights $\bW$ and bias $\mathbf{b}$; $n = 2, \dots, 4$ indexes successive convolutions over atomic neighbourhood shells; the graph operations $\gat_0$, $\gat_1$ are static attention mechanisms combined with a gated recurrent unit $\gru$ (see below), as originally suggested by Xiong and co-workers for predictive modelling~\cite{xiong_pushing_2019,fey19}; and $\lnorm$ is defined by 
\begin{align}
    \lnorm(\bx) = \frac{\bx - \langle \bx \rangle}{\sqrt{\langle \bx^2 \rangle - \langle \bx \rangle^2 }},
\end{align}
where $\langle \dots \rangle$ denotes averaging along the non-batch dimension.

The first graph operator $\gat_0[\bx^\zero,  \be^\zero]$ encapsulates bonding information within the first neighbourhood shell of an atom $a$:
\begin{align}
    \bx^\zero_{\mathcal{N}(a)} &= \sum_{a' \in \mathcal{N}(a)} \gamma_{aa'} \bW' \bx_{a'}^\zero. 
\end{align}
The attention weights are calculated via
\begin{align}
    \bx_{aa'} &= \lrelu\left(\bW \left( \bx_{a'}^\zero || \be_{aa'}^\zero \right) \right),  \\
    z_{aa'} &= \lrelu\left( \bc_1^t \bx_a^\zero + \bc_2^t \bx_{aa'} \right),  \\
    \gamma_{aa'} &= \frac{\exp\left(z_{aa'}\right)}{\sum_{a'' \in \mathcal{N}(a)} \exp \left(z_{aa''}\right)}.  
\end{align}
\begin{figure*}[hbt]
\includegraphics[width=1.0\textwidth]{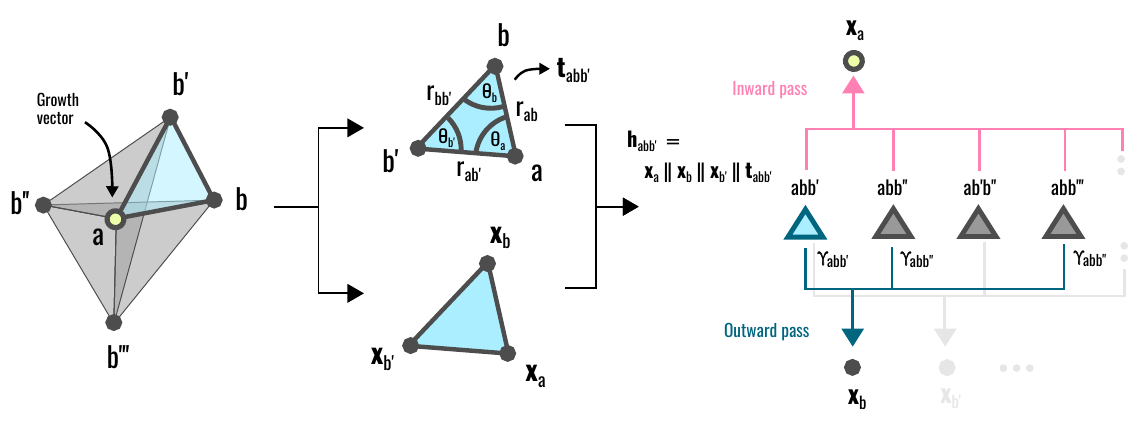}
\caption{ {\footnotesize \textbf{Triangular attention mechanism.} Triplet construction (left), featurization of triangular (hyper-)nodes describing atom triplets $(a,b,b')$ (center), and outward and inward attentive message passing from atom triplets to individual atoms (right). The central growth vector atom is denoted by $a$, environment atoms by $b,b'$ etc.  } }
\label{fig:si_hyper}
\end{figure*}
With bonding information absorbed into the node vectors, the second graph operator $\gat_1[\bx^\zero]$ implements convolutions along the (now unlabeled) covalent edges: 
\begin{align}
    \bx_a^\zero{}' &= \gamma_{aa} \bW \bx_a^\zero + \sum_{a' \in \mathcal{N}(a)} \gamma_{aa'} \bW \bx_{a'}^\zero 
\end{align}
The attention weights are again defined statically:
\begin{align}
    z_{aa'} &= \lrelu\left( \bc^t \left(\bW \bx^\zero_a || \bW \bx^\zero_{a'}\right) \right),   \\
    \gamma_{aa'} &= \frac{\exp\left(z_{aa'}\right)}{\sum_{a'' \in \mathcal{N}(a) \cup \{ a \}} \exp \left(z_{aa''}\right)}.
\end{align}

A gated recurrent unit $\gru(\bh^\none, \bx^\nzero)$ is used to update the atom node vectors $\bx$ based on the current node embeddings and the output of the graph convolution operators~\cite{paszke19}:
\begin{align}
    \br &= \sigmoid\left(\linear\left(\bh^\none\right) + \linear\left(\bx^\nzero\right)\right), \nonumber \\
    \bs &= \sigmoid\left(\linear\left(\bh^\none\right) + \linear\left(\bx^\nzero\right)\right), \nonumber \\
    \bt &= \tanh\left(\linear\left(\bh^\none\right) + r \circ \linear\left(\bx^\nzero\right)\right), \nonumber \\
    \bx^\none &= (1 - \bs) \circ \bt + \bs \circ \bx^\nzero. \nonumber
\end{align}

The output of this topological atom encoder (Eq.~\ref{eq:atomenc_start}-\ref{eq:atomenc_end}) are $d$-dimensional atom embeddings that are further processed into 2D growth and motif vectors:
\begin{align}
    \bx_\mu^\zero &= \bx_{2\dd} \ \mathrm{with} \ \mu \in \{0, 1\}, \\
    \bx_\mu^\kone &= \splus\left(\linear\left(\bx_\mu^\kzero\right)\right), \\
    \bx_\mu &= \linear\left(\bx_\mu^{(2)}\right), \\
    \bu_{a,2} &= \bx_{a,0} || \bx_{a,1}, \\
    \bv_{i,2} &= \bx_{a,1} || \bx_{a,0},
\end{align}
where $k = 1, \dots, 3$ is an iteration index; $\mu \in \{0,1\}$ indicates two coexisting representations, which are ultimately concatenated so as to obtain a symmetric scalar product that is suited to predict links between a ligand-like compound and motif-like moiety, but also potentially between two ligand-like inputs:
\begin{align}
    \alpha_2(\bv_i; \bu_{a}) = \sigmoid\left( \frac{\bv_{i,2} \cdot \bu_{a,2}}{2 \sqrt{d}} \right).
\end{align}

\section{3D model}
\label{sec:si_3d}

The 3D model augments the 2D posterior with an additional likelihood factor that accounts for 3D structural and chemical propensities. Given a growth vector $a$, motif vectors are sampled from the full PQR posterior,
\begin{align}
    \bv_{i} &\sim \frac{f(\bv_i)}{\sum_j f(\bv_j) } \times \frac{\alpha^{2\dd}(\bv_{i,2}; \bu_{a,2})}{1 - \alpha^{2\dd}(\bv_{i,2}; \bu_{a,2})} \times \nonumber \\
    &\ \ \times \frac{\alpha^{3\dd}(\bv_{i,3}; \bu_{a,3})}{1 - \alpha^{3\dd}(\bv_{i,3}; \bu_{a,3})}. 
\end{align}
Even though various other implementations are conceivable, the architecture used here to represent the likelihood score function $\alpha_3(\bv_{i,3};\bu_{a,3})$ is based on a simple hypergraph mechanism that performs attentive convolutions over 3D atom triplets.\\

\textbf{Input features.}
The 2D molecular graph is featurized in the same way as described above for the 2D graph-convolutional architecture (Tabs.~\ref{tab:atomfeature} and~\ref{tab:bondfeature}). In addition to atom nodes and covalent bond edges, triangular hypernodes are generated within a spherical volume around each growth vector (see {\it Methods} of the main text). These atom triplets are featurized using basic geometric features as summarized in Tab.~\ref{tab:hyperfeature}. Convolutions over a local neighbourhood hypergraph, consisting of atomic nodes $b$ within a spherical volume around a growth vector $a$, as well as atom triplets $(a,b,b')$ within that same volume, are used to construct the 3D context vector $\bu_{a,3}$ on which the distribution over motifs $\bv_{i}$ is conditioned.\\

\begin{table}[h]
\centering
    \begin{tabular}{l l c}
        Feature & Description & Dim. \\
        \hline
         Node type & $b = b'$ vs $b \neq b'$ & 1 \\
         $r_{ab}$ & Distance encoding (RBF) & $\lceil r_\mathrm{max}/\sigma \rceil+1$ \\
         $r_{ab'}$ & Distance encoding (RBF) & $\lceil r_\mathrm{max}/\sigma \rceil+1$ \\
         $r_{bb'}$ & Distance encoding (RBF) & $\lceil r_\mathrm{max}/\sigma \rceil+1$ \\
         $\cos(\theta_{a})$, $\sin(\theta_{a})$ & Angular phase & 2 \\
         $\cos(\theta_{b})$, $\sin(\theta_{b})$ & Angular phase & 2 \\
         $\cos(\theta_{b'})$, $\sin(\theta_{b'})$ & Angular phase & 2 \\
         \hline
    \end{tabular}
    \caption{\textbf{Hypernode features.} Attributes describing atom triplets $(a,b,b')$ used by the 3D growth-vector embedding network. The growth-vector atom corresponds to atom index $a$. Triplets with $b \neq b'$ are accounted for twice, as $(a,b,b')$ and $(a,b',b)$ to guarantee permutational invariance. Here we used equispaced Gaussian radial basis functions (RBFs) for the distance encoding with width $\sigma=1\unit{\angstrom}$, and cutoff $r_\mathrm{max} = 7.5\unit{\angstrom}$. }
    \label{tab:hyperfeature}
\end{table}

\textbf{Hypergraph architecture.}
Topological embeddings are generated using the $\encoder$ sequence defined in Eqs.~\ref{eq:atomenc_start} to~\ref{eq:atomenc_end}:
\begin{align}
    \bx_{2\dd} = \encoder\left[\bx, \be\right]
\end{align}
An environment index is constructed around every growth vector atom using a spherical cutoff, with the atoms of those spherical neighbourhoods described by $\xenv^\zero$, obtained from the 2D embeddings $\bx_{2\dd}$ using slicing along the batch dimension. 3D context vectors are constructed using the following set of convolutions:
\begin{align}
    \xenv^\zero{}' &= \transfer\left(\xenv^\zero\right), \\
    \xenv^\one &= \tat_\mathrm{outward} \left[ \xenv^\zero{}', \bt \right], \\
    \xenv^\two &= \tat_\mathrm{inward} \left[ \xenv^\one, \bt \right], \\
    \xenv^\two{}' &= \transfer\left(\xenv^\two\right), \\
    \bx_{3\dd} &= \bW \xenv^\two{}'.
\end{align}
Here $\transfer$ is a residual transfer block, defined by
\begin{align}
    \bx' &= \elu\left(\linear\left(\bx\right)\right), \\
    \tilde{\bx}' &= \lnorm\left(\bx'\right), \\
    \bx_\mathrm{out} &= \elu\left(\bx + \linear\left(\tilde{\bx}'\right)\right).
\end{align}
$\tat_\mathrm{outward}$ and $\tat_\mathrm{inward}$ are hypergraph operators, corresponding to an {\it outward} and {\it inward} pass over atom triplets, as shown schematically in Fig.~\ref{fig:si_hyper}. Specifically, $\tat_\mathrm{outward}$ transmits information from the ordered triplets $(a,b,b')$ to the outermost vertex $b'$ via:
\begin{align}
    \bx_\mu &= \linear(\bx) \ \mathrm{with} \ \mu \in \{0,1,2\}, \\
    \bh_{abb'} &= \bx_{a,0}\, ||\, \bx_{b,1}\, ||\, \bx_{b',2}\, ||\, \bt_{abb'}, \\
    \bh' &= \hgat_\mathrm{outward}\left[ \bx, \bh, \bw \right], \label{eq:hga_out}\\
    \bx_\mathrm{out} &= \elu(\bx + \bh').
\end{align}
$\hgat$ is the attention mechanism that controls this information flow, relying on three inputs: the 2D atom embeddings $\bx$, triangular hypernode embeddings $\bh$, and hypernode-to-atom attention priors $\bw$:
\begin{align}
    \bx_{b'} &= \sum_{ab \in \mathcal{H}(b')} \tilde{\gamma}_{abb'} \bW \bh_{abb'}, \\
    \bh_{b',abb'} &= \lrelu\left(\bW' \bx_{b'} + \bW'' \bh_{abb'}\right), \\
    z_{abb'} &= \lrelu\left(\bc^t \bh_{b',abb'}\right), \\
    \gamma_{abb'} &= \frac{\exp\left(z_{abb'}\right)}{\sum_{ \bar{a}\bar{b} \in \mathcal{N}(b') } \exp\left(z_{\bar{a}\bar{b}b'}\right)}, \\
    \tilde{\gamma}_{abb'} &= \frac{\gamma_{abb'} w_{abb'}}{\sum_{ \bar{a}\bar{b} \in \mathcal{H}(b')} \gamma_{\bar{a}\bar{b}b'} w_{\bar{a}\bar{b}b'} }.
\end{align}
$\mathcal{H}(b')$ denotes all pairs of atoms $(a,b)$ that form a triangular hypernode $(a,b,b')$ with atom $b'$ as the terminal vertex. The attention priors $w_{abb'}$ impose smoothness on the attention weights as atoms leave the spherical neighbourhood shell, and furthermore discount {\it remote} over {\it close} triplets (consistent with the Jacobian $1/r^2$ of the 3D space in which we are operating):
\begin{align}
    w_{abb'} &= f_\mathrm{cut}(r_{ab}) f_\mathrm{cut}(r_{ab'}) g(r_{ab}) g(r_{ab'}), \\
        f_\mathrm{cut}(r) &= \left\{ \begin{array}{cl}
         1 & \mathrm{if}\ r \leq r_\mathrm{cut} - \Delta,  \\
         0 & \mathrm{if} \ r \geq r_\mathrm{cut},  \\
         \frac{1 + \cos\left(\pi  \frac{r - r_\mathrm{cut} + \Delta}{\Delta} \right)}{2} & \mathrm{else}, \\ 
    \end{array} \right. \\
    g(r_{ab}) &= \frac{1 - \omega(r_{ab})}{r_{ab'}^2} + \omega(r_{ab}) w_0, \\
    \omega(r) &= \sigmoid\left(- \beta ( r^2 - \rho^2 ) \right).
\end{align}
The hyperparameters of this prior are: the transition width of the cutoff function $\Delta = 0.5\unit{\angstrom}$; limiting weight factor $w_0 = 1$; radial pivot $\rho = r_\mathrm{cut} - 2 \Delta$; and decay constant $\beta = 0.1$.

The inward pass $\tat_\mathrm{inward}$ inverts the information flow towards the central vertex $a$ (i.e., the growth vector). $\tat_\mathrm{inward}$ resembles $\tat_\mathrm{outward}$ above, with, however, Eq.~\ref{eq:hga_out} replaced by
\begin{align}
    \bh' &= \hgat_\mathrm{inward}\left[ \bx, \bh, \bw \right].
\end{align}
$\hgat_\mathrm{inward}\left[ \bx, \bh, \bw \right]$ closely resembles $\hgat_\mathrm{outward}$, with index relationships subtly rearranged according to:
\begin{align}
    \bx_{a} &= \sum_{bb' \in \mathcal{H}(a)} \tilde{\gamma}_{abb'} \bW \bh_{abb'}, \\
    \bh_{a,abb'} &= \lrelu\left(\bW \bx_{a} + \bW' \bh_{abb'}\right), \\
    z_{abb'} &= \lrelu\left(\bc^t \bh_{a,abb'}\right), \\
    \gamma_{abb'} &= \frac{\exp\left(z_{abb'}\right)}{\sum_{ \bar{b}\bar{b}' \in \mathcal{N}(a) } \exp\left(z_{a\bar{b}\bar{b}'}\right)}, \\
    \tilde{\gamma}_{abb'} &= \frac{\gamma_{abb'} w_{abb'}}{\sum_{ \bar{b}\bar{b}' \in \mathcal{H}(a)} \gamma_{a\bar{b}\bar{b}'} w_{a\bar{b}\bar{b}'} },
\end{align}
where the sum over $\mathcal{H}(a)$ runs over all tuples $(b,b')$ that form triplets $(a,b,b')$ with the growth vector $a$.\\

\textbf{Motif encoding.}
The motif encodings $\bv_i$ featuring in the 3D likelihood factor are again based on the 2D atom encoder sequence from Eqs.~\ref{eq:atomenc_start}-\ref{eq:atomenc_end}, with weights shared across the growth and motif embedder networks:
\begin{align}
    \bx_{2\dd} = \encoder\left[\bx, \be\right].
\end{align}
From $\bx_{2\dd}$, we obtain $\bx_{2\dd}^\mathrm{vec}$ as the subset of 2D vectors that correspond to the linker atoms on the motifs; and $\bx_{2\dd}^\mathrm{env}$, that comprises all 2D atom vectors associated with the motif as a whole. Convolutions placed thereon give rise to the final vector that combines the local information reflected by $\bx_{2\dd}^\mathrm{vec}$ with a global motif vector $\bx^\mathrm{env}$ derived from $\bx_{2\dd}^\mathrm{env}$ using attentive pooling:
\begin{align}
    \bx^\mathrm{vec} &= \transfer(\bx_{2\dd}^\mathrm{vec}), \\
    \bx^\mathrm{env} &= \reduce\left[\transfer(\bx_{2\dd}^\mathrm{env})\right], \\
    \bx_{2\dd}^{\mathrm{motif}} &= \linear\left(\bx^{\mathrm{vec}} + \bx^{\mathrm{env}}\right).
\end{align}
Here $\transfer$ is a residual transfer block as defined above; $\reduce$ is an attentive pooling operator consisting of:
\begin{align}
    \bx_{m}^\mathrm{pool} &= \sum_{j \in \mathcal{M}(m)} \bx_{2\dd, mj}^\mathrm{env}, \\
    \Delta \bx_{m}^\mathrm{pool} &= \sum_{j \in \mathcal{M}(m)} \gamma_{mj} \bW \bx_{m}^\mathrm{pool}, \\
    \bx_{m}^\mathrm{pool} &= \elu\left(\bx_{m}^\mathrm{pool} + \Delta \bx_{m}^\mathrm{pool}\right), \\
    \bx_{m}^\mathrm{env} &= \lnorm\left( \bx_{m}^\mathrm{pool} \right),
\end{align}
with $\{ j \in \mathcal{M}(m) \}$ comprising all atoms $j$ participating in motif $m$ (not just the linker atom). The attention weights are evaluated statically:
\begin{align}
    \bh_{mj} &= \lrelu\left(\bW_0 \bx_{mj} + \bW_1 \bx_{m}^{\mathrm{pool}}\right) \\
    z_{mj} &= \lrelu\left(\bc^t \bh_{mj}\right) \\
    \gamma_{mj} &= \frac{\exp(z_{mj})}{\sum_{j \in \mathcal{M}(m)} \exp(z_{mj})}
\end{align}
\\

\textbf{3D likelihood factor.}
We identify the 3D growth vector $\bu_{a,3}$ and motif vector $\bv_{i,3}$ with the following tensor outputs of the hypergraph and graph architectures above:
\begin{align}
    \bu_{a,3} &\equiv \bx_{a,3\dd}, \\
    \bv_{i,3} &\equiv \bx_{a,2\dd}^\mathrm{motif}.
\end{align}
Note therefore that $\bv_{i,3}$ does not incorporate any 3D information, and that the subscript is used only to differentiate it from the motif vector $\bv_{i,2}$ on which the 2D factor $q$ is based. The likelihood score function $\alpha_3$ that gives rise to the 3D likelihood factor $r = \alpha_3 / 1 - \alpha_3$ is, finally, obtained from the activated inner product
\begin{align}
    \alpha_3(\bv_i; \bu_{a}) = \sigmoid\left(\frac{\bv_{i,3} \cdot \bu_{a,3}}{\sqrt{d}} \right).
\end{align}


\section{Entropic shift} 
\label{sec:si_entropy}

\begin{figure}[t]
\includegraphics[width=1.0\columnwidth]{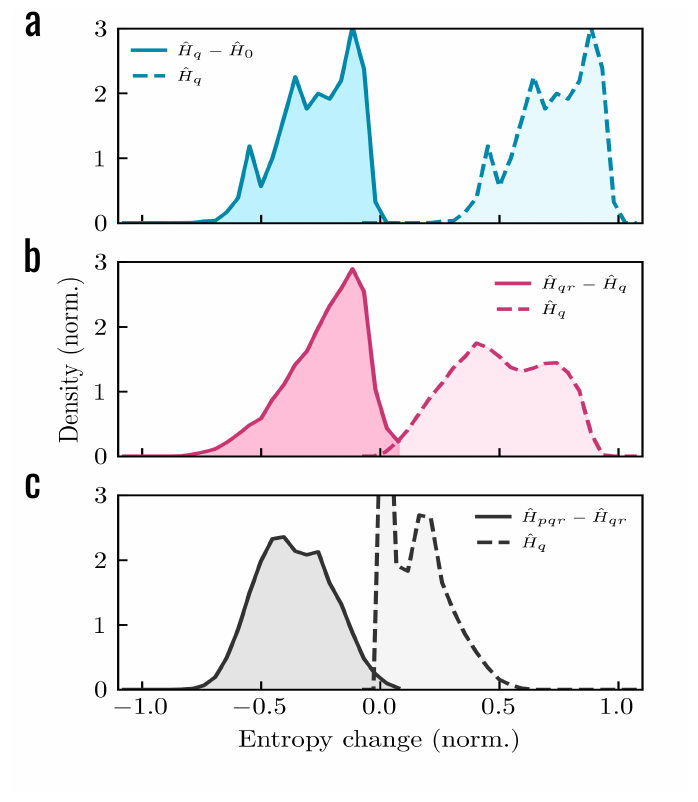}
\caption{ {\footnotesize \textbf{Sequential information in the $\mathbf{QRP}$ model.} Effect of conditioning on the (a) 2D context, (b) 3D context and (c) 1D vocabulary, as quantified by the entropy (solid lines) and entropy change (dashed lines) of the posterior. The normalization applied to the entropy is such that $\hat{H}_0 = 1$ corresponds to the entropy of the uniform distribution over the vocabulary. } }
\label{fig:si_entropy}
\end{figure}

Beyond measuring enrichment, we can estimate the relative information gain achieved by incorporating the 1D, 2D and 3D context by tracking how the entropy of the posterior changes in the QRP sequence (i.e., when conditioning first on the 2D, then 3D, then 1D context). To this end, we evaluate a normalized Shannon entropy $\hat{H}$ over the vocabulary $\voc$,
\begin{align}
    \hat{H}[p_a] = - \frac{1}{\ \log | \voc |} \sum_{\bv_i \in \voc} p_{i|a} \log p_{i|a}, 
\end{align}
where $|\voc|$ is the size of the vocabulary. With the chosen normalization, a uniform prior will result in the maximal possible $\hat{H}_0(a) = 1$. The likelihoods $p_{i|a}$ are given by, initially, the 2D likelihood factors $q_{i|a}$ (or, more precisely, the corresponding normalized probability density); then the combined 2D/3D likelihood factors $q_{i|a} \times r_{i|a}$; and finally the full posterior $p_{i|a} \times q_{i|a} \times r_{i|a}$. 

We denote the corresponding entropies by $\hat{H}_q$, $\hat{H}_{qr}$, $\hat{H}_{pqr}$, respectively, distributions of which are shown in Fig.~\ref{fig:si_entropy}a-c (dashed lines). The same figure also reflects the entropy changes $\hat{H}_q - \hat{H}_0$, $\hat{H}_{qr} - \hat{H}_q$ and $\hat{H}_{pqr} - \hat{H}_{qr}$ (solid lines). Strongly lowering entropy, the 1D and 2D likelihood factors play the largest role in guiding the generative model towards a particular set of motifs, reflecting fragment sociability~\cite{stdenis21} and synthetic accessibility. The effect of the 3D factor $r$ is sizeable but still less pronounced than of $p$ and $q$. This is partially the result of the 2D model occasionally constraining the probability density to a small set of chemically and sterically more or less equivalent choices -- but also a consequence of the fact that the 3D -- as opposed to the 2D -- model most likely still operates relatively far from its theoretical peak performance.\\

\begin{figure*}[hbt]
\includegraphics[width=1.0\textwidth]{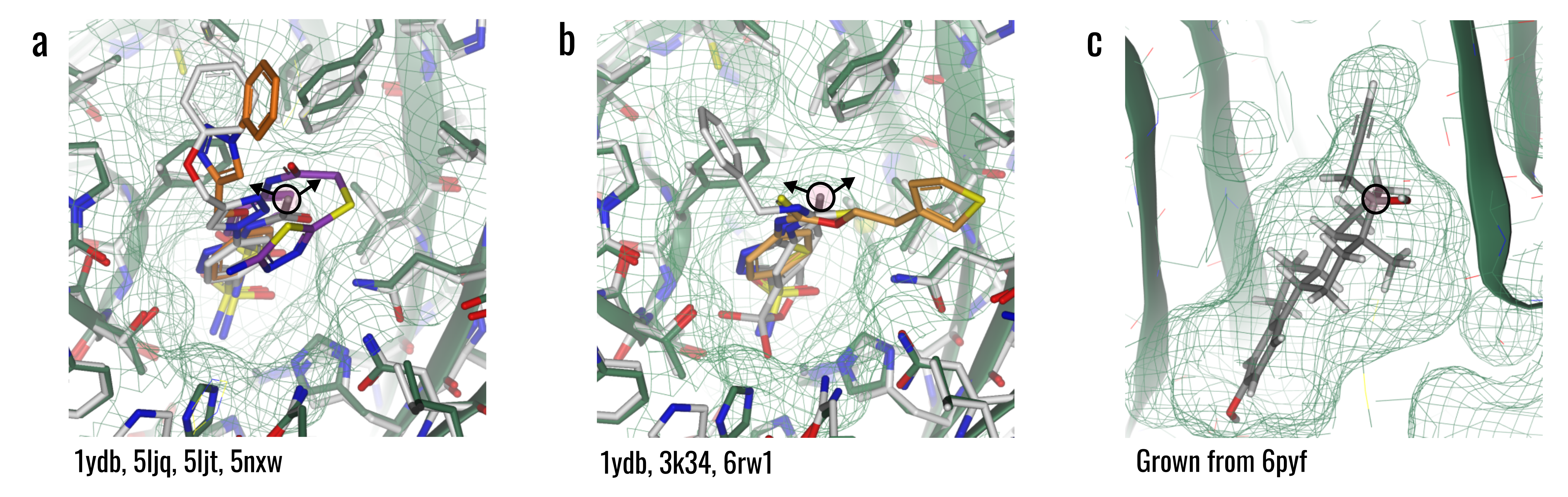}
\caption{ {\footnotesize \textbf{Motif precedents and template-based docking.} (a) Potential precedents for observation of pyrazole and triazole in the neighbourhood of the growth vector of 1ydb (black circle with arrows). (b) Thione motifs in 3k34 and 6rw1 in proximity to the 1ydb growth vector. (c) Template-based docking of a carbyne motif CC$\equiv$C attached to the 6pyf growth vector. } }
\label{fig:si_elaboration}
\end{figure*}

\section{Elaboration examples (Fig.~5)}
\label{sec:si_elaboration}

Validating the output of generative models using computational routes only is typically challenging. In the examples provided in Fig.~5 of the main text, the experimental SAR was recovered in three out of four cases. The lower-rank suggestions are, however, not trivially correct. In example (a: aminopeptidase) the halogen motifs are consistent with an SAR observed by Helgren {\it et al$.$}~\cite{example_3vh9_alt}. In example (b: yellow lupin), the top-ranked bromo motif is consistent with the groud-truth iodo, with otherwise only fluoro achieving significant amplification. 

In example (d: hCA II), substitution of thiophene (rank 1) with other aromatic five-membered rings (ranks 2 and 4) appears broadly plausible. Pyrazole and triazole are, however, distinct from thiophene regarding their electrostatic and lipophilic properties. We have therefore scanned the PDB for precedents of these motifs, with three examples (Fig.~S3a) overlayed with the reference structure 1ydb -- thus indicating that there is some support for introducing aromatic nitrogen into the region of the growth vector. The thiol motifs (ranks 3 and 5) are, however, precedented only as thiones (Fig.~S3b). Even though the 2D bonding model can assign a double bond to the link between the thiol and the growth vector (thus converting thiol to thione), the resulting thio-aldehyde is a known PAINS structure~\cite{pains} and therefore to be avoided.

The 3D ranking failed to recover the butyl motif from example (c: SHBG). Carbyne motifs, notably, the three-atom vector CC$\equiv$C can be easily incorporated into the environment, resulting in an adequate fit as demonstrated by template-based docking (Fig.~S3c).

\end{bibunit}


\end{document}